\documentclass[runningheads]{llncs}

\usepackage{eccv}

\usepackage{eccvabbrv}
\usepackage{graphicx}
\usepackage[percent]{overpic}
\usepackage{xcolor}
\usepackage{caption}
\usepackage{subcaption}
\usepackage[export]{adjustbox}
\usepackage{booktabs}
\usepackage[table]{xcolor}
\usepackage{longtable}
\usepackage{svg}
\usepackage{bbm}
\usepackage{placeins}

\usepackage{amsmath}

\newcommand{\repeatcommand}[2]{%
  \ifnum#1>0
    #2%
    \repeatcommand{\numexpr#1-1\relax}{#2}%
  \fi
}

\usepackage[accsupp]{axessibility}  %

\usepackage{hyperref}

\usepackage{orcidlink}

\makeatletter
\renewcommand*{\@fnsymbol}[1]{\ensuremath{\ifcase#1\or *\or \dagger\or \ddagger\or
    \mathsection\or \mathparagraph\or \|\or **\or \dagger\dagger
    \or \ddagger\ddagger \else\@ctrerr\fi}}
\makeatother

\begin{document}

\title{InterEdit: Navigating Text-Guided 3D Dyadic Human Motion Editing} 

\titlerunning{InterEdit: Navigating Text-Guided 3D Dyadic Human Motion Editing}

\author{Yebin Yang\inst{1} \orcidlink{0009-0002-5770-6744}
\and Di Wen\inst{1} \orcidlink{0009-0000-1693-7912}
\and Lei Qi\inst{1} \orcidlink{0009-0005-9001-2623}
\and Weitong Kong\inst{1}
\and Junwei Zheng\inst{1} \orcidlink{0009-0005-4390-3044}
\and Ruiping Liu\inst{1} \orcidlink{0000-0001-5245-2277}
\and Yufan Chen\inst{1} \orcidlink{0009-0008-3670-4567}
\and Chengzhi Wu\inst{1} \orcidlink{0000-0003-2186-3748}
\and Kailun Yang\inst{2} \orcidlink{0000-0002-1090-667X} 
\and Yuqian Fu\inst{3} \orcidlink{0000-0002-0412-5500}
\and Danda Pani Paudel\inst{4} \orcidlink{0000-0002-1739-1867}
\and Luc Van Gool\inst{4} \orcidlink{0000-0002-3445-5711}
\and Kunyu Peng\inst{1,4,}\thanks{Correspondence: kunyu.peng@kit.edu} \orcidlink{0000-0002-5419-9292}
}

\authorrunning{Y.~Yang~\textit{et al.}}

\institute{Karlsruhe Institute of Technology, Germany
\and Hunan University, China
\and KAUST, Saudi Arabia
\and INSAIT, Sofia University ``St. Kliment Ohridski'', Bulgaria}

\maketitle

\begin{abstract}
Text-guided 3D motion editing has seen success in single-person scenarios, but its extension to multi-person settings is less explored due to limited paired data and the complexity of inter-person interactions. We introduce the task of multi-person 3D motion editing, where a target motion is generated from a source and a text instruction. To support this, we propose \textbf{InterEdit3D}, a new dataset with manual two-person motion change annotations, and a Text-guided Multi-human Motion Editing (TMME) benchmark. We present \textbf{InterEdit}, a synchronized classifier-free conditional diffusion model for TMME. It introduces Semantic-Aware Plan Token Alignment with learnable tokens to capture high-level interaction cues and an Interaction-Aware Frequency Token Alignment strategy using DCT and energy pooling to model periodic motion dynamics. Experiments show that \textbf{InterEdit} improves text-to-motion consistency and edit fidelity, achieving state-of-the-art TMME performance. The dataset and code will be released at \url{https://github.com/YNG916/InterEdit}.
\end{abstract}

\section{Introduction}

Human 3D motion generation~\cite{zhang2025energymogen,zhuo2025infinidreamer,lucas2022posegpt,azadi2023make,zhu2023human,Guo_2022_CVPR,lee2024t2lm,li2025unimotion,liang2024intergen,javedintermask} has seen significant advancements with large-scale datasets and diffusion-based models, enabling realistic behavior synthesis from textual descriptions. However, pure text generation is often insufficient for practical content creation, where animators, game designers, and AI systems require \emph{controlled modifications} of existing motions rather than full synthesis~\cite{li2025simmotionedit}.
Text-guided motion editing~\cite{wang2025fg,hong2025salad,zheng2025diffmesh,athanasiou2024motionfix,guo2025motionlab,wang2025TIMotion,zhang2025kinmo,wang2025unitmge} addresses this by allowing users to modify a source motion based on a text instruction, producing a target motion that changes only the requested parts while preserving other content. This controllable refinement is crucial for iterative design, personalization, and human-AI collaboration.

While text-driven motion editing~\cite{meng2025rethinking,hong2025salad,zheng2025diffmesh,athanasiou2024motionfix,guo2025motionlab} has shown promising results for single-person scenarios, extending it to multi-human interactions remains underexplored. Many real-world human behaviors involve interactions—such as collaboration, competition, and physical contact—that require multiple participants~\cite{peng2025hopadiff,peng2024referring}.
In these cases, motion meaning arises not only from individual movements but from \emph{spatio-temporal coupling}, including synchronization, phase alignment, positioning, role switching, and contact timing. Editing such interactions is crucial for applications like character animation, social robotics, virtual agents, crowd simulation, and training data for embodied intelligence, where fine-grained control over interaction dynamics is essential.
Editing multi-human motion is more challenging than single-actor editing due to the interaction semantics in relative timing and configuration. A small modification to one person can disrupt synchronization or spatial consistency. Unlike generation, editing must stay anchored to the source motion and apply only instruction-relevant changes. This ``change what is requested, preserve the rest'' constraint is harder in interactive settings, where small temporal deviations can alter semantics. Despite its importance, there is no dedicated benchmark for multi-human 3D motion editing.

\begin{figure*}[t!]

    \centering
    \begin{subfigure}{0.6\textwidth}
        \centering
        \includegraphics[width=\linewidth]{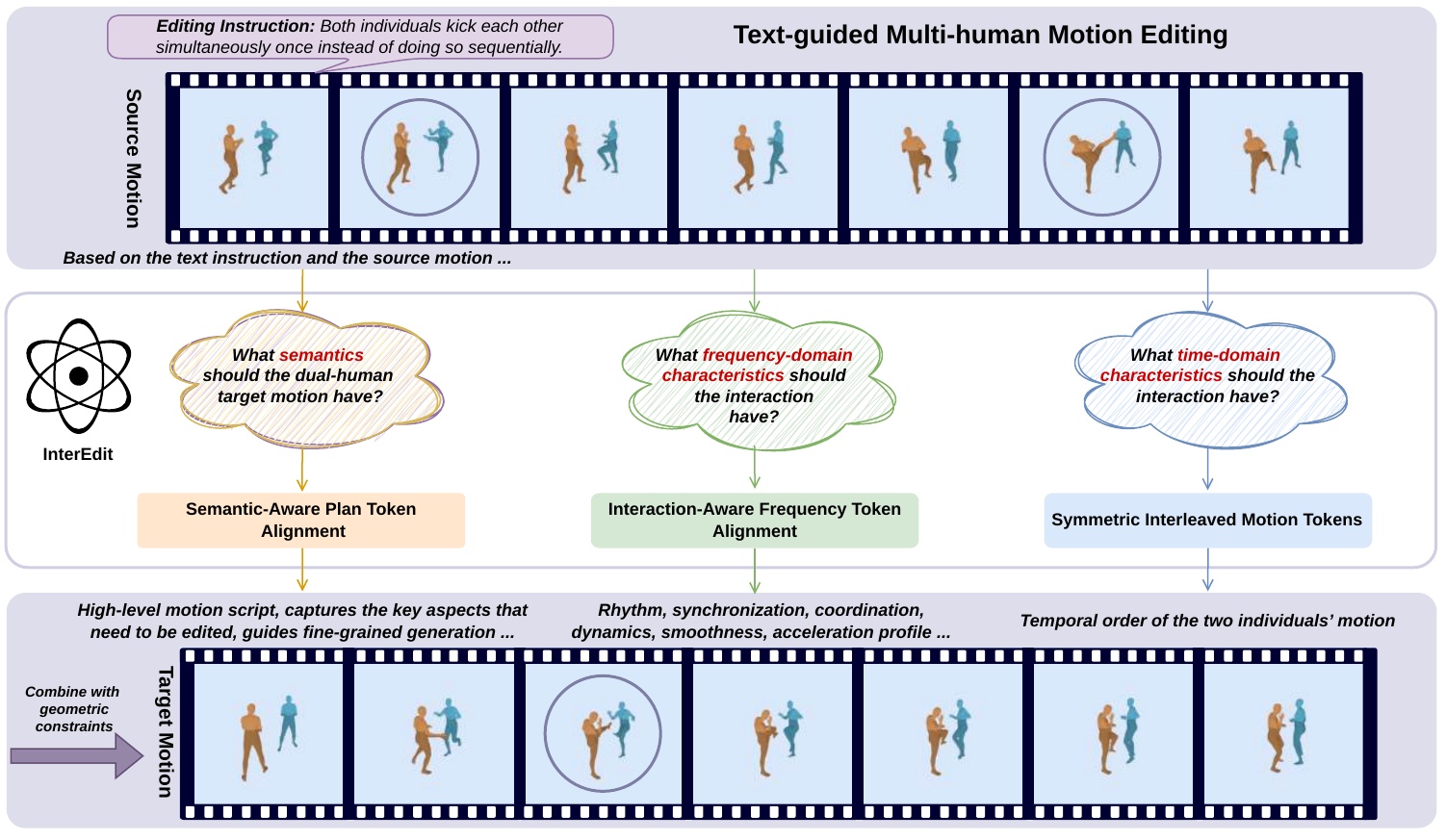}
        \caption{An overview of the task and \textbf{InterEdit}.}
        \label{fig:sub1}
    \end{subfigure}
    \hfill
    \begin{subfigure}{0.39\textwidth}
        \centering
        \includegraphics[width=\linewidth]{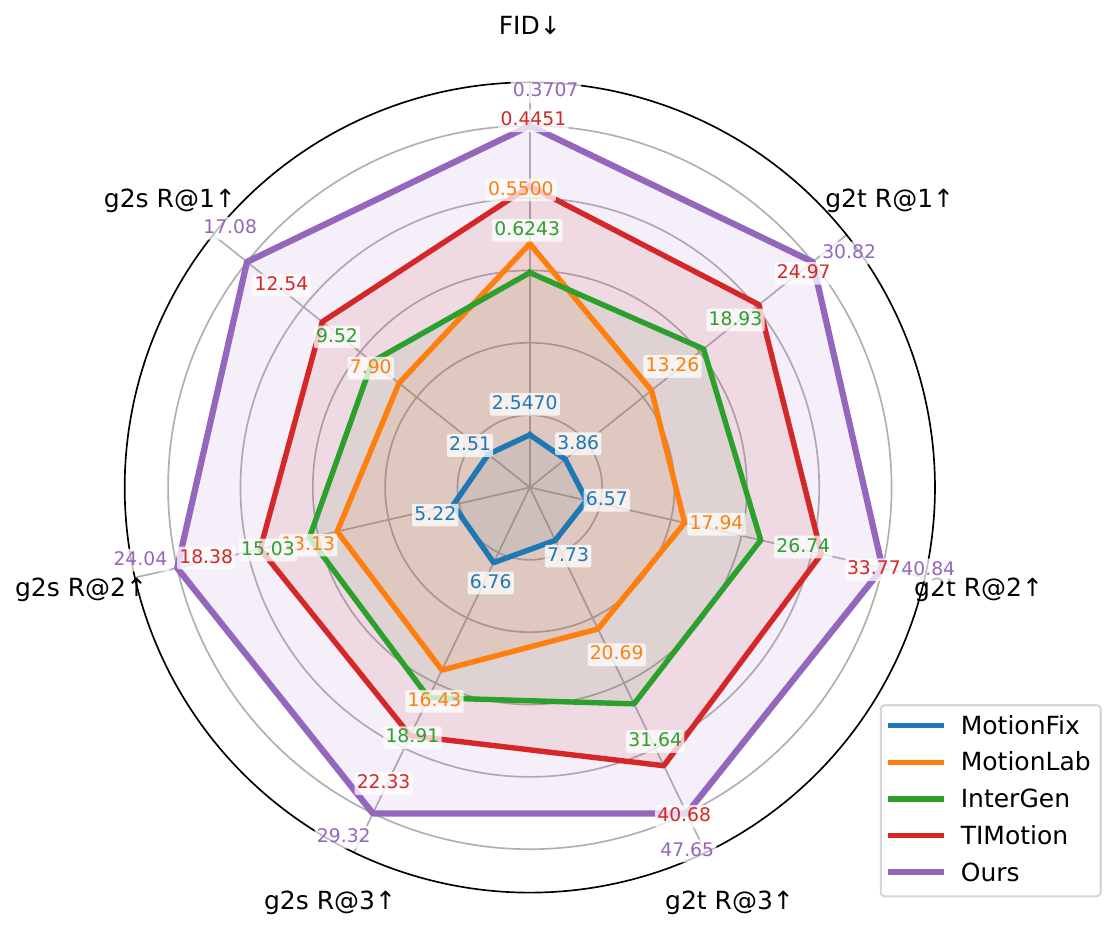}
        \caption{An overview of performances of baselines and our \textbf{InterEdit}.}
        \label{fig:sub2}
    \end{subfigure}
    \caption{An illustration of (a) Text-guided Multi-human 3D Motion Editing (TMME) task and our proposed \textbf{InterEdit} model, and (b) the performances of baselines (\ie, MotionFix~\cite{athanasiou2024motionfix}, MotionLab~\cite{guo2025motionlab}, InterGen~\cite{liang2024intergen}, TIMotion~\cite{wang2025TIMotion}) and our \textbf{InterEdit}.}
    \label{fig:teaser}
\end{figure*}

In this work, we introduce \emph{Text-guided Multi-human Motion Editing} (TMME), a new task that aims to generate a target two-person motion conditioned on both source motion and textual editing instruction (Fig.~\ref{fig:teaser}). This task exposes two major challenges. First, paired source--target--instruction triplets for multi-human motion are scarce, as existing datasets are primarily designed for generation. Second, interaction editing requires precise modeling of coordination patterns such as synchrony versus alternation, tempo variation, spatial approach and separation, and role-dependent responses.

To enable systematic study of this problem, we construct \textbf{InterEdit3D}, the first large-scale two-person motion editing dataset built on top of InterHuman~\cite{liang2024intergen} via a semi-automatic retrieval-and-annotation pipeline. We retrieve motion pairs that share a similar base motion for one individual while differing in interaction structure, and annotate them with editing instructions describing how to transform the source into the target. This results in $5{,}161$ source--target--text triplets emphasizing spatial, temporal, and coordination-level edits, forming a challenging benchmark for interaction-aware motion editing.

We propose \textbf{InterEdit}, a classifier-free conditional diffusion framework specifically designed for dual-person motion editing. To determine what to modify while preserving source consistency, we introduce \emph{Semantic-Aware Plan Token Alignment}: learnable plan tokens aligned with a pretrained motion teacher embedding of the target motion, providing semantic-level guidance and ensuring that the edits align with the high-level intent expressed in the text instruction.
To preserve temporal coupling and coordination, we further introduce \emph{Interaction-Aware Frequency Token Alignment}.
\emph{Interaction-Aware Frequency Token Alignment} applies the Discrete Cosine Transformation (DCT) to average and difference interaction signals to obtain band-energy descriptors, which are mapped into frequency control tokens.
These tokens are supervised to regress the target band-energy profiles, encouraging edits that respect interaction rhythm, synchrony, and other coupling dynamics.
By combining Semantic-Aware Plan Token Alignment and Interaction-Aware Frequency Token Alignment, \textbf{InterEdit} effectively captures both high-level editing intent and fine-grained interaction dynamics, which are crucial for generating realistic and semantically accurate multi-human motion edits.

Our main contributions are summarized as follows:
\begin{itemize}
\item We introduce the Text-guided Multi-human Motion Editing (TMME) benchmark, the first benchmark for multi-person motion editing, with $4$ single-person editing and multi-person generation baselines.
\item We present \textbf{InterEdit3D}, the first large-scale multi-person motion editing dataset containing $5,161$ source--target--text triplets, created through a scalable retrieval pipeline.
\item We propose \textbf{InterEdit}, a multi-human conditional diffusion framework incorporating semantic-aware plan token alignment and interaction-aware frequency token alignment, ensuring precise and temporally synchronized motion edits and delivering state-of-the-art TMME performance.
\end{itemize}

\section{Related Work}

\noindent\textbf{3D Human Motion Generation.}
\underline{\textit{Single-Person Human Motion Generation:}}
Text-driven single-person motion generation~\cite{uchida2025mola,zhang2025energymogen,zhuo2025infinidreamer,lucas2022posegpt,azadi2023make,zhu2023human,Guo_2022_CVPR,lee2024t2lm,li2025unimotion} has rapidly advanced with diffusion models becoming the dominant paradigm.
Representative works include MDM~\cite{tevet2023mdm}, VAE~\cite{bie2022hit,yan2018mt,petrovich2021action,petrovich2022temos}, auto-regressive normalization network~\cite{henter2020moglow}, GAN~\cite{yan2017skeleton,xu2023actformer,barsoum2018hp,ghosh2021synthesis,amballa2025ls}, motion diffusion~\cite{kapon2024mas,ren2023diffusion,zhang2022motiondiffuse}, latent-space diffusion~\cite{kong2023priority,chen2023mld,zhang2025energymogen} for improved efficiency and fidelity.
Beyond direct text conditioning, retrieval-augmented generation further improves text--motion alignment and diversity by incorporating nearest-neighbor motion candidates~\cite{zhang2023remodiffuse}.
Large-scale datasets and stronger text encoders (\eg, CLIP~\cite{radford2021clip}) have also contributed to better language understanding and more expressive motion synthesis.
\underline{\textit{Multi-Person Motion Generation:}}
Generating human-human interactions is more challenging than single-person synthesis due to mutual dependencies and spatio-temporal coupling~\cite{liang2024intergen,javedintermask,wang2025TIMotion}. Multi-person motion generation is enabled by the proposal of two-person motion datasets~\cite{liang2024intergen,javedintermask}.
InterGen~\cite{liang2024intergen} introduces a large-scale interaction dataset and a diffusion model for joint denoising of both participants. 
In2IN~\cite{ruiz2024in2in} enhances controllability by leveraging individual-level information, while InterControl~\cite{wang2024intercontrol} explores structured control for interaction synthesis. 
InterMask~\cite{javedintermask} improves spatio-temporal consistency via masked prediction, and TIMotion~\cite{wang2025TIMotion} emphasizes temporal and interactive dynamics. 
Recent methods like InterMoE~\cite{wang2025intermoeindividualspecific3dhuman} and HINT~\cite{liu2026hinthierarchicalinteractionmodeling} focus on individual-specific and hierarchical modeling, respectively. However, existing works primarily focus on 3D multi-human generation based on text alone.

\noindent\textbf{Text-Guided Single-Person 3D Motion Editing.}
Previous work on 3D human motion editing focused on spatial or temporal constraints~\cite{lee1999hierarchical,gleicher2001motion,gleicher1997motion}, with recent deep learning methods balancing instruction adherence and source preservation~\cite{meng2025rethinking,hong2025salad,zheng2025diffmesh,athanasiou2024motionfix,guo2025motionlab,wang2025TIMotion,zhang2025kinmo,wang2025unitmge,hu2023motion}. MotionFix~\cite{athanasiou2024motionfix} uses diffusion for text-driven single-person editing, but extending this to dual-person interactions is challenging due to the need to preserve realism and spatio-temporal coupling. 
Similarity-aware methods like SimMotionEdit~\cite{li2025simmotionedit} limit deviation from the source, but multi-person editing is still underexplored.
Unlike prior work on human-to-human interaction generation~\cite{liang2024intergen,wang2025TIMotion,ruiz2024in2in} and single-person editing~\cite{athanasiou2024motionfix}, we explore \emph{multi-person motion editing}, conditioning on a multi-person source motion and instruction. We propose semantic-aware plan token alignment and interaction-aware frequency token alignment to ensure precise motion edits while preserving temporal coordination and minimizing drift in multi-human interactions.

\section{Dataset and Benchmark}

\noindent\textbf{Dataset: InterEdit3D.}
We present \textbf{InterEdit3D}, a multi-human 3D motion editing dataset composed of (source motion, target motion, edit text) triplets. It is built upon InterHuman~\cite{liang2024intergen}, a large-scale 3D human–human interaction dataset featuring diverse two-person interactions with natural language descriptions, \eg, daily activities (greeting, handshaking, object passing) and professional interactions (martial arts, dancing).
To construct editing pairs, we retrieve motion pairs that share similar motion backbones but differ in interaction semantics via motion-to-motion retrieval. Each pair is further annotated with an instruction that describes how to transform the source motion into the target motion, forming high-quality supervision for multi-person motion editing.

\noindent\underline{Dataset Comparison.}
Table~\ref{tab:dataset_compare} compares our dataset with others across \textit{text annotation}, \textit{interaction}, and \textit{editing}. Classical datasets like KIT-ML~\cite{Plappert_2016}, BABEL~\cite{punnakkal2021babelbodiesactionbehavior}, and HumanML3D~\cite{Guo_2022_CVPR} focus on single-person motions but lack multi-person interactions or editing pairs. PoseFix~\cite{PoseFix_2023_ICCV} targets pose correction, while MotionFix~\cite{athanasiou2024motionfix} enables single-person motion editing but lacks interactions. Interaction datasets like InterHuman~\cite{liang2024intergen} and Inter-X~\cite{xu2024inter} focus on two-person motion generation but don't offer editing triplets.
Our dataset uniquely combines text, interaction, and editing, offering paired source-target motions with edit instructions for two-person interactions. This enables evaluation of models that preserve consistency and follow instructions without disrupting interactions.

\begin{table}[t]
\centering
\setlength{\tabcolsep}{8pt}
\begin{tabular}{lcccccc}
\hline
Dataset & Text & Interactive & Editing & Motions & Vocab.\\
\hline
KIT-ML~\cite{Plappert_2016}      & \checkmark & -- & -- & 3{,}911   & 1{,}623 \\
BABEL~\cite{punnakkal2021babelbodiesactionbehavior}    & \checkmark & -- & -- & 10{,}881  & 1{,}347 \\
HumanML3D~\cite{Guo_2022_CVPR}     & \checkmark & -- & -- & 14{,}616  & 5{,}371 \\
PoseFix~\cite{PoseFix_2023_ICCV}    & \checkmark & -- & \checkmark & 6{,}157 (pairs) & 1{,}068 \\
MotionFix~\cite{athanasiou2024motionfix} & \checkmark & -- & \checkmark & 6{,}730 (pairs) & 1{,}479 \\
\hline
InterHuman~\cite{liang2024intergen} & \checkmark & \checkmark & -- & 6{,}022 & 5{,}656 \\
Inter-X~\cite{xu2024inter}        & \checkmark & \checkmark & -- & 11{,}388 & 3{,}467 \\
\hline
\textbf{Ours}                      & \checkmark & \checkmark & \checkmark &
\textbf{5{,}161 (pairs)} & \textbf{1754} \\
\hline
\end{tabular}
\caption{Comparison with representative existing datasets. ``Editing'' indicates whether the dataset provides source--target pairs and edit instructions.}
\label{tab:dataset_compare}
\end{table}

\noindent\underline{Data Collection and Annotation.}
We convert the motion format provided by InterHuman~\cite{liang2024intergen} into AMASS features~\cite{mahmood2019amass} and encode them using the pretrained TMR~\cite{petrovich2023tmrtexttomotionretrievalusing} motion encoder, which provides semantically meaningful embeddings via contrastive motion–text alignment. Motion-to-motion retrieval is performed in this latent space.
We retrieve the top-2 nearest neighbors (cosine similarity), forming source--target candidates that share a similar base motion for one individual but differ in interaction semantics, enabling ``edit the change, preserve the rest.''
Annotators write an \textit{edit instruction} describing how to transform the source into the target. Subtle or unclear pairs are discarded.
In total, we obtain $5{,}161$ triplets, split into train/validation/test sets with an $80\%$/$10\%$/$10\%$ ratio. We perform interaction-level disjoint split by manually checking to ensure no overlapping interaction identities among different sets. The dataset is annotated by $8$ individuals, and a cross-checking process is performed to ensure consistency in the annotations. Samples with significantly divergent annotations are discarded.
Overall, \textbf{InterEdit3D} emphasizes relative spatial control, temporal ordering, semantic action changes, and body-part-aware interaction edits, forming a realistic and challenging benchmark for dual-person motion editing.

\noindent\textbf{Baselines.}
We compare with 4 representative methods from the two closest settings: single-person motion editing models MotionFix~\cite{athanasiou2024motionfix} and MotionLab~\cite{guo2025motionlab}, and multi-human motion generation models InterGen~\cite{liang2024intergen} and TIMotion~\cite{wang2025TIMotion}. Since no prior method is specifically designed for text-guided dyadic motion editing with source--target--instruction triplets, we adapt these methods and retrain them on \textbf{InterEdit3D}. All baselines use the same motion representation, data splits, and evaluation metrics as \textbf{InterEdit}.
\noindent\underline{Single-Person Editing Baselines.}
We concatenate both individuals’ motion features along the feature dimension, and train the models to predict the edited two-person target motion from the source motion and editing instruction.
\noindent\underline{Multi-Human Motion Generation Baselines.} We preserve their original interaction modeling architectures and inject the source motion as an additional condition via AdaLN, enabling motion generation conditioned on both the source motion and instruction.

\section{Methodology}
\begin{figure*}[t!]
    \includegraphics[width=\textwidth]{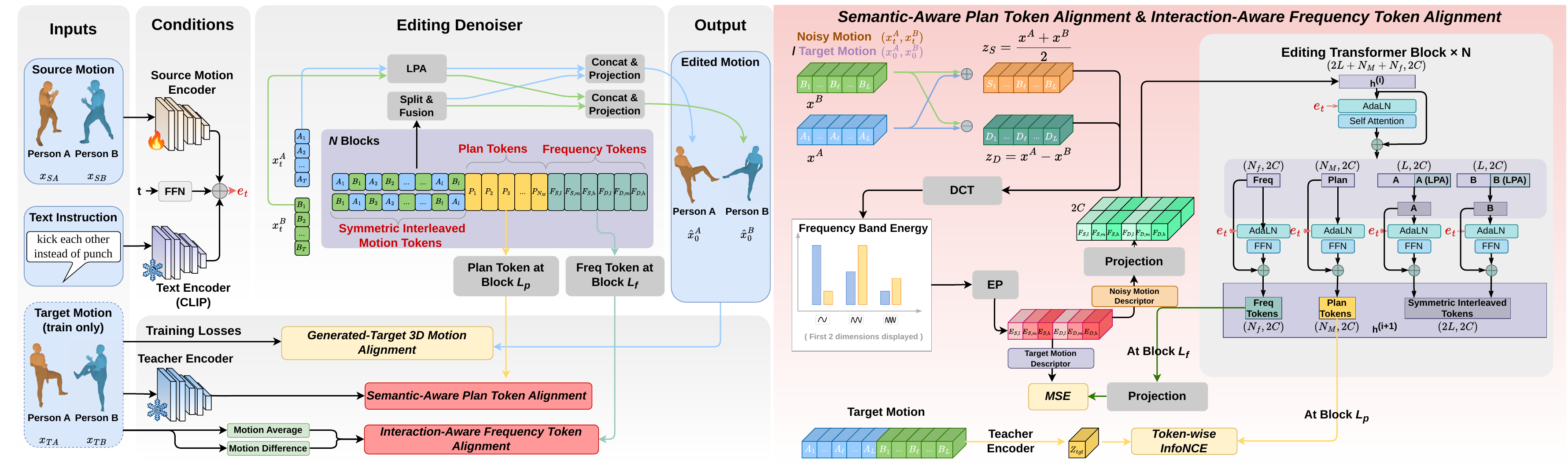}
\caption{Overview of the proposed \textbf{InterEdit} framework.
Given a two-person motion and an editing instruction, \textbf{InterEdit} uses a conditional diffusion backbone with symmetric interleaved motion tokens.
It introduces (i) \emph{Semantic-Aware Plan Token Alignment} for high-level editing guidance via a motion-teacher embedding, and (ii) \emph{Interaction-Aware Frequency Token Alignment} using DCT-based band-energy descriptors to regulate interaction dynamics.}
\vskip-3ex
    \label{fig:main}
\end{figure*}

As shown in Fig.~\ref{fig:main}, we formulate two-person motion editing as a conditional diffusion model, where the denoiser is conditioned on the source motion and the text instruction, further guided by semantic-aware plan token alignment and interaction-aware frequency token alignment.

\subsection{Multi-Human Motion Representation}
\label{sec:repr}

We follow the non-canonical motion representation commonly used for two-person 3D motion data following InterHuman~\cite{liang2024intergen}.
For each person $p\in\{A,B\}$ at frame $\ell$, the motion state is as Eq.~\ref{eq:1}.
\begin{equation}
\label{eq:1}
\mathbf{x}^{p}_{\ell}=\big[\mathbf{j}^{p}_{g,\ell},\ \mathbf{v}^{p}_{g,\ell},\ \mathbf{r}^{p}_{\ell},\ \mathbf{c}^{p}_{\ell}\big]\in\mathbb{R}^{d_m},
\end{equation}
where
$\mathbf{j}^{p}_{g,\ell}\in\mathbb{R}^{3N_j}$ and $\mathbf{v}^{p}_{g,\ell}\in\mathbb{R}^{3N_j}$ denote global joint positions and velocities in the world frame, $\mathbf{r}^{p}_\ell\in\mathbb{R}^{6N_j}$ is the 6D local joint rotation representation in the root frame, $N_j$ denotes the number of joints per person, and $\mathbf{c}^{p}_\ell\in\mathbb{R}^{4}$ is the binary foot--ground contact.
A two-person motion sequence is denoted as Eq.~\ref{eq:2}.
\begin{equation}
\label{eq:2}
\mathbf{x}_{1:L}=\big(\mathbf{x}^{A}_{1:L},\mathbf{x}^{B}_{1:L}\big)\in\mathbb{R}^{L\times 2d_m},
\end{equation}
where $L$ denotes the motion length (number of frames). Given a two-person source motion sequence $\mathbf{x}^s_{1:L}$ and an edit instruction text $\mathbf{y}$, our goal is to generate an edited two-person target motion $\hat{\mathbf{x}}_{1:L}$ that follows the instruction while preserving source-consistent content.

\subsection{Multi-Human Conditional Diffusion for Motion Editing}
\label{sec:diff}

We formulate multi-human motion editing as conditional generation with a diffusion model
$p_\theta(\mathbf{x}_0\mid \mathbf{x}^s, \mathbf{y})$, where $\mathbf{x}^s$ is the source motion sequence,
$\mathbf{y}$ is the editing instruction, and $\mathbf{x}_0$ denotes the clean target motion sequence.
Let $\mathbf{x}_t$ denote the noisy version of $\mathbf{x}_0$ at diffusion step $t\in\{1,\dots,T\}$.
Following a noise schedule $\{\beta_t\}_{t=1}^{T}$, the forward noising process is as Eq.~\ref{eq:3}.
\begin{equation}
\label{eq:3}
q(\mathbf{x}_t\mid \mathbf{x}_{t-1})=\mathcal{N}\!\left(\sqrt{1-\beta_t}\,\mathbf{x}_{t-1},\beta_t\mathbf{I}\right),
\qquad
q(\mathbf{x}_t\mid \mathbf{x}_0)=\mathcal{N}\!\left(\sqrt{\bar{\alpha}_t}\,\mathbf{x}_0,(1-\bar{\alpha}_t)\mathbf{I}\right),
\end{equation}
where $\alpha_t=1-\beta_t$ and $\bar{\alpha}_t=\prod_{i=1}^{t}\alpha_i$.
Equivalently, we can sample according to Eq.~\ref{eq:4}.
\begin{equation}
\label{eq:4}
\mathbf{x}_t=\sqrt{\bar{\alpha}_t}\,\mathbf{x}_0+\sqrt{1-\bar{\alpha}_t}\,\boldsymbol{\epsilon},
\quad \boldsymbol{\epsilon}\sim\mathcal{N}(\mathbf{0},\mathbf{I}).
\end{equation}

\noindent\textbf{Denoising Model and Conditioning.}
We use a transformer denoiser $\mathcal{D}_\theta$ with \textsc{Start\_X} parameterization, i.e., it directly predicts the clean motion $\hat{\mathbf{x}}_0$ as shown in Eq.~\ref{eq:5} (rather than the added noise $\boldsymbol{\epsilon}$)~\cite{ho2020ddpm}; this prediction is used to form the reverse-step distribution (and is compatible with DDIM~\cite{song2022ddim} sampling).
\begin{equation}
\label{eq:5}
\hat{\mathbf{x}}_0 = \mathcal{D}_\theta(\mathbf{x}_t,t;\ \mathbf{c}_{\text{text}},\mathbf{c}_{\text{src}}),
\end{equation}
where $\theta$ denotes trainable parameters, $\mathbf{c}_{\text{text}}$ is the instruction embedding extracted through a frozen CLIP textual encoder~\cite{radford2021clip} via $\mathbf{c}_{\text{text}} = \text{CLIP}(\mathbf{y})$, and
$\mathbf{c}_{\text{src}}$ is the source-motion embedding.
We encode the source motion $\mathbf{x}^s$ with a lightweight learnable Transformer source-motion encoder to obtain $\mathbf{c}_{\text{src}}$; for this encoding we remove the last 4 foot-contact channels per person and concatenate two persons along the feature dimension, then project it into a latent space, prepend a learnable query (CLS) token, and apply positional encoding followed by a multi-layer Transformer encoder; the source-motion embedding is taken from the output of the query token.
Both conditions are injected into the denoiser via AdaLN modulation as Eq.~\ref{eq:6}.
\begin{equation}
\label{eq:6}
\mathbf{e}_t = \mathrm{EmbedTime}(t) + W_{\text{text}}\mathbf{c}_{\text{text}} + W_{\text{src}}\mathbf{c}_{\text{src}},
\end{equation}
where $\mathrm{EmbedTime}(\cdot)$ is a sinusoidal timestep embedding followed by an MLP projection, and $W_{\text{text}}$, $W_{\text{src}}$ are linear projections.

\noindent\textbf{Symmetric Interleaved Token Aggregation.}
To model temporal-order influence and role switching in two-person interactions, we follow TIMotion~\cite{wang2025TIMotion} and construct a symmetric interleaved sequence rather than directly concatenating the two persons into a single token stream. 
Let $\mathbf{x}^A_c,\mathbf{x}^B_c\in\mathbb{R}^{L\times C}$ be the motion token sequences for two persons,  where $L$ is the length of the sequence and $C$ is the dimension of motion embedding. We build a causal interleaving $\mathbf{x}_{\mathrm{cii}}\in\mathbb{R}^{2L\times C}$ and its role-swapped counterpart $\mathbf{x}_{\mathrm{sym}}\in\mathbb{R}^{2L\times C}$ as Eq.~\ref{eq:7}.
\begin{equation}
\label{eq:7}
\begin{aligned}
\mathbf{x}_{\mathrm{cii}}(2\ell{-}1) &= \mathbf{x}^A_c(\ell), &
\mathbf{x}_{\mathrm{cii}}(2\ell) &= \mathbf{x}^B_c(\ell),\\
\mathbf{x}_{\mathrm{sym}}(2\ell{-}1) &= \mathbf{x}^B_c(\ell), &
\mathbf{x}_{\mathrm{sym}}(2\ell) &= \mathbf{x}_c^A(\ell).
\end{aligned}
\end{equation}
We then feed the concatenation result $\mathbf{x}_{inter}=\mathrm{Concat}(\mathbf{x}_{\mathrm{cii}},\mathbf{x}_{\mathrm{sym}})\in\mathbb{R}^{2L\times 2C}$ into a transformer. Given an output $\hat{\mathbf{x}}_{inter}\in\mathbb{R}^{2L\times 2C}$, we split channels $\hat{\mathbf{x}}_{({inter},1)}=\hat{\mathbf{x}}_{inter}[:,{:}C]$, $\hat{\mathbf{x}}_{({inter},2)}=\hat{\mathbf{x}}_{inter}[:,C{:}]$ and de-interleave odd/even indices to recover per-person streams, then merge the two role views by element-wise sum to obtain $\hat{\mathbf{x}}_c^{A,g},\hat{\mathbf{x}}_c^{B,g}\in\mathbb{R}^{L\times C}$. We further apply a lightweight Localized Pattern Amplification (LPA)~\cite{wang2025TIMotion} branch to refine short-range temporal patterns for each person. Given the per-person motion token sequence, $\mathbf{x}_c^p\in\mathbb{R}^{L\times C}$, we modulate it with the condition embedding $\mathbf{e}_t$ via AdaLN and extract local features with a small 1D convolutional stack as Eq.~\ref{eq:8}.
\begin{equation}
\label{eq:8}
\hat{\mathbf{x}}^{p}_c = \mathbf{x}^{p}_c + \mathrm{Conv}_{1}\!\left(\mathrm{AdaLN}(\mathrm{Conv}_{3}\!\left(\mathrm{AdaLN}(\mathbf{x}^{p}_c,\mathbf{e}_t)\right),\mathbf{e}_t)\right),\qquad
\end{equation}
where $\mathrm{Conv}_{k}$ denotes a 1D convolution with kernel size $k$.
We then fuse global and local features by channel-wise concatenation followed by a linear projection back to $C$ channels as Eq.~\ref{eq:lpa_final}.
\begin{equation}
\mathbf{x}^{p,f}_c=
\mathrm{Linear}\!\left(\mathrm{Concat}\big(\hat{\mathbf{x}}_c^{p,g},\hat{\mathbf{x}}^{p}_c\big)\right),\quad
p\in\{A,B\}.
\label{eq:lpa_final}
\end{equation}
Then the final outputs are fed into the next transformer block until the final decoder layer to produce the denoiser prediction.

\noindent\textbf{Training objective.}
With \textsc{Start\_X} parameterization, we train the denoiser $\mathcal{D}_{\theta}$ by minimizing the reconstruction error of $\mathbf{x}_0$ as Eq.~\ref{eq:10}.
\begin{equation}
\label{eq:10}
\mathcal{L}_{\text{diff}}
=\mathbb{E}_{t,\mathbf{x}_0,\boldsymbol{\epsilon}}
\Big[\big\|\mathbf{x}_0 - \mathcal{D}_\theta(\mathbf{x}_t,t;\mathbf{c}_{\text{text}},\mathbf{c}_{\text{src}})\big\|_2^2\Big].
\end{equation}
\subsection{Synchronized Classifier-Free Guidance for Editing}
\label{sec:cfg}
We introduce Synchronized Classifier-Free Guidance (SCFG) for controllable editing. During training, we randomly drop conditioning with probability \(p_{\text{scfg}}\) by setting both \(\mathbf{c}_{\text{text}}\) and \(\mathbf{c}_{\text{src}}\) to zeros in a synchronized way. This results in an unconditional branch. At inference, we combine the conditional and unconditional predictions as Eq.~\ref{eq:11}.
\begin{equation}
\label{eq:11}
\hat{\mathbf{x}}_{0,\mathrm{scfg}}
=
\gamma\,\mathcal{D}_\theta(\mathbf{x}_t,t;\mathbf{c}_{\text{text}},\mathbf{c}_{\text{src}})
+(1-\gamma)\,\mathcal{D}_\theta(\mathbf{x}_t,t;\mathbf{0},\mathbf{0}),
\end{equation}
which is equivalent to
$\hat{\mathbf{x}}_{0,\mathrm{scfg}} = \mathcal{D}_\theta(\cdot;\mathbf{0},\mathbf{0}) + \gamma\big(\mathcal{D}_\theta(\cdot;\mathbf{c}_{\text{text}},\mathbf{c}_{\text{src}})-\mathcal{D}_\theta(\cdot;\mathbf{0},\mathbf{0})\big)$.
Here $\gamma$ is the guidance scale.
Dropping both conditions avoids leakage (source/text bias) and yields a cleaner guidance direction.

\subsection{Semantic-Aware Plan Token Alignment}
\label{sec:plan}

Multi-human editing requires identifying \emph{what to change} while preserving non-edited content.
We introduce learnable \textbf{plan tokens} to provide semantic-level guidance.

\noindent\textbf{Plan Tokens as Control Tokens.}
We append $N_M$ learnable plan tokens $\mathbf{P}\in\mathbb{R}^{N_M\times 2C}$ to the denoiser token sequence.
Through self-attention, motion tokens can query $\mathbf{P}$ and receive global guidance during denoising.
At transformer block $L_p$, we project plan tokens into a semantic space as Eq.~\ref{eq:12}.
\begin{equation}
\label{eq:12}
\hat{\mathbf{z}}^{(k)} = g_p\!\left(\mathbf{P}^{(L_p)}_{k}\right)\in\mathbb{R}^{C},\quad k=\left[1,\dots,N_M\right],
\end{equation}
where $g_p$ is a combination of layer normalization and linear projection.

\noindent\textbf{Teacher Target Embedding.}
We use a frozen motion teacher encoder $f_T(\cdot)$ (trained contrastively on InterHuman~\cite{liang2024intergen} motion–text pairs) to extract a target semantic embedding from the ground-truth target motion as Eq.~\ref{eq:13}.
\begin{equation}
\label{eq:13}
\mathbf{z}_{\text{tgt}} = f_T(\mathbf{x}_0)\in\mathbb{R}^{C}.
\end{equation}
This teacher embedding provides a compact semantic target for aligning plan tokens.

\noindent\textbf{Plan Token Alignment Loss.}
We apply a token-wise InfoNCE objective with $\mathbf{z}_{\text{tgt}}$ as the positive target and in-batch negatives ($n$ indexes targets in the mini-batch).
Let $\tilde{\mathbf{z}}_{\text{tgt}}=\mathrm{norm}(\mathbf{z}_{\text{tgt}})$ and $\tilde{\mathbf{z}}^{(k)}=\mathrm{norm}(\hat{\mathbf{z}}^{(k)})$.
The plan loss is shown in Eq.~\ref{eq:14}.
\begin{equation}
\label{eq:14}
\mathcal{L}_{\text{plan}}
=
\frac{1}{N_M}\sum_{k=1}^{N_M}
\left[
-\log
\frac{\exp\left((\tilde{\mathbf{z}}^{(k)})^\top\tilde{\mathbf{z}}_{\text{tgt}}/\tau\right)}
{\sum_{n}\exp\left((\tilde{\mathbf{z}}^{(k)})^\top\tilde{\mathbf{z}}_{\text{tgt}}^{(n)}/\tau\right)}
\right],
\end{equation}
where $\tau$ is the temperature.
We also experimented with cosine and MSE variants, but InfoNCE~\cite{oord2019infonce} provides the best overall trade-off.

\subsection{Interaction-Aware Frequency Token Alignment}
\label{sec:freq}
Interaction correctness is highly sensitive to temporal coupling (synchrony vs.\ alternation, phase alignment, contact timing).
We propose Interaction-Aware Frequency Token Alignment, which leverages Discrete Cosine Transformation (DCT) and energy pooling to capture and regularize the frequency dynamics of interaction signals, ensuring precise synchronization and coordination.

\noindent\textbf{Interaction Signals: Average and Difference.}
We exclude foot-contact channels (last 4 dims per person) and denote the remaining feature dimension as $d_f$.
For a two-person motion sequence $\mathbf{x}=(\mathbf{x}^A,\mathbf{x}^B)$, we construct two interaction-aware sequences as Eq.~\ref{eq:15}.
\begin{equation}
\label{eq:15}
\mathbf{z}_S=\frac{\mathbf{x}^{A}+\mathbf{x}^{B}}{2}\in\mathbb{R}^{L\times d_f},
\qquad
\mathbf{z}_D=\mathbf{x}^{A}-\mathbf{x}^{B}\in\mathbb{R}^{L\times d_f},
\end{equation}
where $\mathbf{z}_S$ captures shared or synchronized components, $\mathbf{z}_D$ captures relative or oppositional components.

\noindent\textbf{Discrete Cosine Transformation (DCT) and Band-Energy Pooling.}
We apply discrete cosine transformation along the time axis as Eq.~\ref{eq:16}.
\begin{equation}
\label{eq:16}
\mathbf{C}_S=\mathrm{DCT}(\mathbf{z}_S),\qquad \mathbf{C}_D=\mathrm{DCT}(\mathbf{z}_D).
\end{equation}
We compute band-energy descriptors for low/mid/high frequency bins by pooling squared coefficients within each band $b$ as Eq.~\ref{eq:17}. 
\begin{equation}
\label{eq:17}
\mathbf{E}(\mathbf{C};b)=\sqrt{\frac{1}{|b|}\sum_{k\in b}\mathbf{C}[k]^2+\epsilon}\in\mathbb{R}^{d_f},
\end{equation}
where $\epsilon$ is a small constant added for numerical stability. Using normalized cutoffs $(r_{\text{l}},r_{\text{m}},r_{\text{h}})$ to define the three bins (low (l), middle (m), and high (h)), we obtain six band-energy descriptors according to Eq.~\ref{eq:18}.
\begin{equation}
\label{eq:18}
\mathbf{g}(\mathbf{x})=
\Big[
\mathbf{E}(\mathbf{C}_S;\text{l}),\mathbf{E}(\mathbf{C}_S;\text{m}),\mathbf{E}(\mathbf{C}_S;\text{h}),
\mathbf{E}(\mathbf{C}_D;\text{l}),\mathbf{E}(\mathbf{C}_D;\text{m}),\mathbf{E}(\mathbf{C}_D;\text{h})
\Big]\in\mathbb{R}^{6\times d_f},
\end{equation}
where $\mathbf{g}(\mathbf{x})$ is computed from $\mathbf{x}$ through $(\mathbf{z}_S,\mathbf{z}_D)$ construction and DCT.

\noindent\textbf{Interaction-Aware Frequency Tokens and Regression Alignment.}
At each diffusion step, we compute band-energy descriptors from the current noisy motion $\mathbf{x}_t$ and project them into the model token space to form $N_f{=}6$ frequency control tokens according to Eq.~\ref{eq:19}.
\begin{equation}
\label{eq:19}
\mathbf{F}_i=\phi_f^{(i)}\!\left(\mathbf{g}_i(\mathbf{x}_t)\right)\in\mathbb{R}^{2C},\quad i=\left[1,\dots,N_f\right],
\end{equation}
where $\phi_f^{(i)}$ is a lightweight network consisting of a layer normalization and a linear projection layer, and $\mathbf{F}=[\mathbf{F}_1;\dots;\mathbf{F}_{N_f}]$ is appended to the motion token sequence and plan token for joint self-attention.
At transformer block $L_f$, we decode the band-energy descriptors from the corresponding token as Eq.~\ref{eq:20}.
\begin{equation}
\label{eq:20}
\hat{\mathbf{g}}_i = g_f^{(i)}\!\left(\mathbf{F}^{(L_f)}_i\right)\in\mathbb{R}^{d_f},\quad i=\left[1,\dots,N_f\right],
\end{equation}
where $g_f^{(i)}$ is a combination of layer normalization and linear projection head. We minimize a weighted regression loss against the ground-truth target band-energy descriptors computed from the clean target motion $\mathbf{x}_0$ according to Eq.~\ref{eq:21}.
\begin{equation}
\label{eq:21}
\mathcal{L}_{\text{freq}}
=
\frac{1}{N_f}\sum_{i=1}^{N_f} w_i \left\|\hat{\mathbf{g}}_i-\mathbf{g}_i(\mathbf{x}_0)\right\|_2^2,
\end{equation}
where $w_i\ge 0$ is a per-term weight. We down-weight the two high-frequency terms (the ``high'' bins for both $\mathbf{z}_S$ and $\mathbf{z}_D$) with a smaller constant weight to reduce sensitivity to high-frequency noise; for all other bins we set $w_i=1$.
We further apply frequency-token dropout during training: with probability $p_f$, we remove frequency tokens from the denoiser, which regularizes training and helps preserve generation quality.

\subsection{Objective Function}
\label{sec:obj}
\noindent\textbf{Diffusion Reconstruction Loss:}
The \textbf{diffusion reconstruction loss} ($\mathcal{L}_{\text{diff}}$) minimizes the MSE between the predicted motion $\hat{\mathbf{x}}_0$ and the ground-truth motion $\mathbf{x}_0$, ensuring that the denoised motion closely matches the original.

\noindent\textbf{Geometric Losses (Per Person):}
We apply several geometric losses to preserve realism in each person's motion. The \textbf{velocity loss} ($\mathcal{L}_{\text{vel}}$) ensures smooth movement by penalizing the difference between predicted and ground-truth velocities. The \textbf{foot-contact loss} ($\mathcal{L}_{\text{foot}}$) ensures realistic foot placement. The \textbf{bone-length loss} ($\mathcal{L}_{\text{BL}}$) maintains consistent bone lengths by comparing per-bone distances between predicted and ground-truth joint positions.

\noindent\textbf{Interaction Losses (Between Persons):}
For realistic interaction, we introduce interaction losses. The \textbf{masked distance-map loss} ($\mathcal{L}_{\text{DM}}$) focuses on joint distances in contact regions. The \textbf{relative-orientation loss} ($\mathcal{L}_{\text{RO}}$) ensures coherent interaction by penalizing orientation mismatches. The motion objective is shown in Eq.~\ref{eq:22}.

\begin{equation}
\label{eq:22}
\begin{aligned} \mathcal{L}_{\text{motion}} &= \mathcal{L}_{\text{diff}} + \lambda_{\text{vel}}\mathcal{L}_{\text{vel}} + \lambda_{\text{foot}}\mathcal{L}_{\text{foot}} + \lambda_{\text{BL}}\mathcal{L}_{\text{BL}} \\ &\quad + \lambda_{\text{DM}}\mathcal{L}_{\text{DM}} +  \lambda_{\text{RO}}\mathcal{L}_{\text{RO}}. \end{aligned} 
\end{equation}

\noindent\textbf{Auxiliary Alignment Terms:}  
To further improve the motion generation, we add two auxiliary alignment terms. The \textbf{plan token alignment} ($\mathcal{L}_{\text{plan}}$) aligns the learnable plan tokens with the teacher embedding of the ground-truth target, guiding the model to capture and represent the overall motion plan accurately. The \textbf{frequency token alignment} ($\mathcal{L}_{\text{freq}}$) aligns the frequency control tokens with the DCT band-energy descriptors derived from interaction signals, preserving the frequency characteristics of the motion and ensuring that the rhythmic aspects of the movement are preserved. The final training objective is shown in Eq.~\ref{eq:23}.
\begin{equation}
\label{eq:23}
\mathcal{L}_{\text{total}} = \mathcal{L}_{\text{motion}} + \lambda_{p}\mathcal{L}_{\text{plan}} + \lambda_{f}\mathcal{L}_{\text{freq}}.
\end{equation}
In this approach, the CLIP text encoder~\cite{radford2021clip} and the motion teacher are kept frozen, and all other parameters are trained end-to-end.

\section{Experiments}
\subsection{Implementation Details}
We implement \textbf{InterEdit} with $N=5$ transformer blocks, each with 16 attention heads and motion embedding dimension 512. The text encoder is a frozen CLIP ViT-L/14~\cite{radford2021clip}. As the motion teacher, we adopt a frozen contrastively trained motion encoder on InterHuman~\cite{liang2024intergen} to provide semantic target embeddings for plan-token alignment. We use $N_M=16$ plan tokens.
For auxiliary losses, $\lambda_p=0.03$ and $\lambda_f=0.01$, applied at blocks $L_p=3$ and $L_f=5$. High-frequency components are down-weighted by 0.25. Three DCT bands are used with $r_{\text{low}}=0.08$, $r_{\text{mid}}=0.25$, and $r_{\text{high}}=0.35$, and frequency tokens are randomly dropped with probability $p_f=0.04$. Other geometric and interaction losses follow InterGen~\cite{liang2024intergen}.
Training uses 1000 diffusion steps with a cosine schedule. At inference we adopt DDIM sampling (50 steps, $\eta=0$) with classifier-free guidance (drop rate $p_{\text{scfg}}=0.1$, guidance scale $\gamma=3.5$). Optimization uses AdamW~\cite{loshchilov2019decoupledweightdecayregularization} ($\beta_1=0.9$, $\beta_2=0.999$), weight decay $2\times10^{-5}$, and peak learning rate $10^{-4}$ with cosine decay and 10 warm-up epochs.
The full model contains 358.8M parameters including frozen encoders, with 85.0M trainable parameters, close to TIMotion's 81.2M and substantially smaller than InterGen's 251M. We train the model for 1500 epochs with batch size 32 on 8 NVIDIA RTX Pro 6000 Blackwell GPUs in about 4h35m, with about 16GB peak GPU memory. At inference, InterEdit achieves 2.842 samples/s, comparable to TIMotion's 2.709 samples/s.

\subsection{Analysis of the Benchmark}

\noindent\textbf{Metrics.}
We use retrieval-based metrics as primary measures following MotionFix~\cite{athanasiou2024motionfix}. For a 3D multi-human motion editing result, we evaluate: (i) generated-to-target retrieval (g2t) and (ii) generated-to-source retrieval (g2s) in a learned motion embedding space, reporting Recall@K ($K \in {1,2,3}$). We use the InterGen text-to-motion retrieval model~\cite{liang2024intergen} as the feature extractor, where motions are L2-normalized and ranked by cosine similarity against the full test set. g2t measures instruction adherence, while g2s reflects source preservation. A strong editor should achieve high g2t with reasonably high g2s, balancing semantic modification and content preservation. We also report FID in the same embedding space to assess motion realism, measuring the distribution distance between generated and real target motions (lower is better). All methods are evaluated for $20$ independent runs, and we report the mean with 95\% confidence intervals.

\noindent\textbf{Benchmark Results and Baseline Comparison.} 
\begin{table}[t!]
\centering
\resizebox{\linewidth}{!}{
    \begin{tabular}{l c *{6}{c}}
    \toprule
    & & \multicolumn{3}{c}{generated-to-source retrieval (\%) $\uparrow$} & \multicolumn{3}{c}{ \textbf{generated-to-target retrieval (\%) $\uparrow$}} \\
    \cmidrule(lr){3-5} \cmidrule(lr){6-8}
    Method & FID $\downarrow$ & R@1 & R@2 & R@3 & \textbf{R@1} & \textbf{R@2} & \textbf{R@3} \\
    \midrule
    MotionFix~\cite{athanasiou2024motionfix} &
    2.1089$\pm$0.0043 &
    3.10$\pm$0.39 & 7.20$\pm$0.34 & 9.10$\pm$0.49 &
    \cellcolor{gray!15}11.60$\pm$0.52 & \cellcolor{gray!15}17.80$\pm$0.47 & \cellcolor{gray!15}21.10$\pm$0.44 \\
    MotionLab~\cite{guo2025motionlab} &
    0.4284$\pm$0.0121 &
    12.40$\pm$0.69 & 18.90$\pm$0.79 & 23.00$\pm$0.38 &
    \cellcolor{gray!15}18.50$\pm$0.79 & \cellcolor{gray!15}25.60$\pm$0.50 & \cellcolor{gray!15}30.50$\pm$0.44 \\
    InterGen~\cite{liang2024intergen} &
    0.6243$\pm$0.0062 &
    9.52$\pm$0.42 & 15.03$\pm$0.44 & 18.91$\pm$0.47 &
    \cellcolor{gray!15}18.93$\pm$0.60 & \cellcolor{gray!15}26.74$\pm$0.71 & \cellcolor{gray!15}31.64$\pm$0.72 \\
    TIMotion~\cite{wang2025TIMotion} &
    0.4451$\pm$0.0058 &
    12.54$\pm$0.34 & 18.38$\pm$0.39 & 22.33$\pm$0.46 &
    \cellcolor{gray!15}24.97$\pm$0.59 & \cellcolor{gray!15}33.77$\pm$0.56 & \cellcolor{gray!15}40.68$\pm$0.65 \\
    \textbf{Ours} &
    \textbf{0.3707$\pm$0.0029} &
    \textbf{17.08$\pm$0.41} & \textbf{24.04$\pm$0.46} & \textbf{29.32$\pm$0.49} &
    \cellcolor{gray!15}\textbf{30.82$\pm$0.43} & \cellcolor{gray!15}\textbf{40.84$\pm$0.70} & \cellcolor{gray!15}\textbf{47.65$\pm$0.59} \\
    \bottomrule
    \end{tabular}
}
\caption{Quantitative comparison (mean, 95\% CI).}
\label{tab:quantitative_comparison}
\end{table}
\begin{table}[t!]
\centering
\scriptsize
\renewcommand{\arraystretch}{0.90}
\setlength{\tabcolsep}{3pt}
\begin{tabular*}{\linewidth}{@{\extracolsep{\fill}}lccc@{}}
\toprule
Method & R-Precision@1$\uparrow$ & R-Precision@2$\uparrow$ & R-Precision@3$\uparrow$ \\
\midrule
InterGen~\cite{liang2024intergen} &
0.371$\pm$0.010 & 0.515$\pm$0.012 & 0.624$\pm$0.010 \\
TIMotion~\cite{wang2025TIMotion} &
0.491$\pm$0.005 & 0.648$\pm$0.004 & 0.724$\pm$0.004 \\
\textbf{Ours} &
\textbf{0.523$\pm$0.004} & \textbf{0.665$\pm$0.005} & \textbf{0.753$\pm$0.004} \\
\bottomrule
\end{tabular*}
\caption{Generalization on the InterHuman generation benchmark (mean, 95\% CI).}
\label{tab:interhuman_generation}
\end{table}
\begin{table}[t!]
\centering
\resizebox{\linewidth}{!}{
    \begin{tabular}{lc *{6}{c}}
    \toprule
    & & \multicolumn{3}{c}{generated-to-source retrieval (\%) $\uparrow$} & \multicolumn{3}{c}{ \textbf{generated-to-target retrieval (\%) $\uparrow$}} \\
    \cmidrule(lr){3-5} \cmidrule(lr){6-8}
    Method & FID $\downarrow$ & R@1 & R@2 & R@3 & \textbf{R@1} & \textbf{R@2} & \textbf{R@3} \\
    \midrule
    without plan/freq token &
    0.4451$\pm$0.0058 &
    12.54$\pm$0.34 & 18.38$\pm$0.39 & 22.33$\pm$0.46 &
    \cellcolor{gray!15}{24.97$\pm$0.59} & \cellcolor{gray!15}{33.77$\pm$0.56} & \cellcolor{gray!15}{40.68$\pm$0.65} \\
    only plan token($L_p=5$) &
    0.3667$\pm$0.0023 & 
    14.52$\pm$0.36 & 21.58$\pm$0.31 & 25.87$\pm$0.32 & \cellcolor{gray!15}{28.72$\pm$0.49} & \cellcolor{gray!15}{37.92$\pm$0.33} & \cellcolor{gray!15}{43.50$\pm$0.44} \\
    only freq token($L_f=5$) &
    0.3798$\pm$0.0027 & 
    14.24$\pm$0.34 & 20.91$\pm$0.39 & 25.48$\pm$0.40 &
    \cellcolor{gray!15}{28.75$\pm$0.47} & \cellcolor{gray!15}{38.46$\pm$0.48} & \cellcolor{gray!15}{44.05$\pm$0.43} \\
    with plan\&freq token &
    0.3707$\pm$0.0029 &
    17.08$\pm$0.41 & 24.04$\pm$0.46 & 29.32$\pm$0.49 &
    \cellcolor{gray!15}{30.82$\pm$0.43} & \cellcolor{gray!15}{40.84$\pm$0.70} & \cellcolor{gray!15}{47.65$\pm$0.59} \\
    
    \bottomrule
    \end{tabular}
}
\caption{Ablation of module component (mean, 95\% CI).}
\label{tab:module_ablation}
\end{table}
\begin{table}[t!]
\centering
\resizebox{\linewidth}{!}{
    \begin{tabular}{l c *{6}{c}}
    \toprule
    & & \multicolumn{3}{c}{generated-to-source retrieval (\%) $\uparrow$} & \multicolumn{3}{c}{ \textbf{generated-to-target retrieval (\%) $\uparrow$}} \\
    \cmidrule(lr){3-5} \cmidrule(lr){6-8}
    Method & FID $\downarrow$ & R@1 & R@2 & R@3 & \textbf{R@1} & \textbf{R@2} & \textbf{R@3} \\
    \midrule
    $p_f=0.05$ &
    0.3477$\pm$0.0020 &
    16.11$\pm$0.37 & 23.14$\pm$0.50 & 27.69$\pm$0.53 &
    \cellcolor{gray!15}29.10$\pm$0.61 & \cellcolor{gray!15}39.46$\pm$0.65 & \cellcolor{gray!15}45.25$\pm$0.62 \\
    $p_f=0.04$  &
    0.3655$\pm$0.0031 &
    16.33$\pm$0.45 & 22.99$\pm$0.38 & 27.07$\pm$0.38 &
    \cellcolor{gray!15}30.51$\pm$0.45 & \cellcolor{gray!15}40.10$\pm$0.32 & \cellcolor{gray!15}45.97$\pm$0.43 \\
    $p_f=0.03$  &
    0.3693$\pm$0.0029 &
    15.20$\pm$0.31 & 21.57$\pm$0.39 & 25.97$\pm$0.38 &
    \cellcolor{gray!15}29.70$\pm$0.44 & \cellcolor{gray!15}40.25$\pm$0.49 & \cellcolor{gray!15}46.48$\pm$0.44 \\
    \bottomrule
    \end{tabular}
}
\caption{Ablation of frequency tokens dropout rate ($L_p=5$, $L_f=5$, mean, 95\% CI).}
\label{tab:dropf}
\end{table}
Table~\ref{tab:quantitative_comparison} presents quantitative comparisons on our dataset with four adapted baselines: single-person editing (MotionFix~\cite{athanasiou2024motionfix}, MotionLab~\cite{guo2025motionlab}) and two-person generation (InterGen~\cite{liang2024intergen}, TIMotion~\cite{wang2025TIMotion}). Our method outperforms all baselines in g2t and g2s, and achieves the lowest FID, demonstrating superior instruction adherence, edit fidelity, and motion realism.
Compared to the strongest baseline TIMotion~\cite{wang2025TIMotion}, InterEdit improves g2t R@1/2/3 by +5.85, +7.07, and +6.97, improves g2s R@1/2/3 by +4.54, +5.66, and +6.99, and reduces FID by 16.7\%, indicating better edits without disrupting interaction coupling.
The adapted single-person editors lag clearly behind in g2t, highlighting the importance of interaction-aware modeling for dual-person motions. Two-person generation baselines achieve better g2t than adapted single-person editors, but g2s remains limited, since they are not designed to balance instruction following with source preservation, which can lead to global drift or disrupted interaction consistency. In contrast, \textbf{InterEdit} combines semantic-aware plan token alignment and interaction-aware frequency token alignment to better capture editing intent and temporal coordination, producing more faithful and realistic dyadic motion edits.

\noindent\textbf{Human Preference Evaluation.}
To complement automatic metrics, we conduct a human preference study against TIMotion. InterEdit is preferred by 75.5\%, 78.5\%, 71.0\%, and 81.0\% on overall preference, instruction adherence, source preservation, and interaction realism, respectively, further confirming its advantages in faithful and realistic interaction editing.

\noindent\textbf{Generalization.}
Beyond our TMME benchmark in Table~\ref{tab:quantitative_comparison}, we further evaluate InterEdit on the standard InterHuman two-person generation benchmark, where motions are generated from text only. 
As shown in Table~\ref{tab:interhuman_generation}, InterEdit achieves the best R-Precision across all ranks, suggesting that our token alignment design also benefits two-person motion modeling beyond editing.

\subsection{Analysis of the Ablation Study}

\noindent\textbf{Effect of the two main modules.} 
Table~\ref{tab:module_ablation} evaluates our core modules, including semantic-aware latent representation alignment and interaction-oriented frequency alignment, individually and jointly. 
Removing both modules (``without plan/freq token'') results in the weakest performance across g2s/g2t retrieval and FID, indicating that the base diffusion model alone struggles to preserve source content and capture target semantics.
Introducing only plan tokens improves both retrieval metrics and motion quality, showing that semantic alignment provides a global editing signal. Using only frequency tokens enhances retrieval and maintains competitive FID, emphasizing the importance of regularizing interaction dynamics. 
Combining both plan and frequency tokens yields the best results, highlighting their complementary roles: plan tokens guide what to edit semantically, while frequency tokens stabilize interaction dynamics.

\noindent\textbf{Randomly dropping frequency tokens during training.} 
Table~\ref{tab:dropf} studies random frequency-token dropout ($p_f$) where we stochastically disable the frequency-token pathway during training. 
We observe that a moderate drop rate provides the best balance: it improves retrieval while keeping FID favorable. 
Intuitively, this stochastic dropping regularizes training by preventing the model from over-relying on the frequency tokens, encouraging robust generation quality even when frequency guidance is partially absent. 
Too small a drop rate may lead to over-dependence, while too large a drop rate weakens the intended frequency alignment signal; both cases yield inferior overall trade-offs compared to the moderate setting. More ablation studies are shown in the supplementary.
\begin{figure*}[h]
 \centering
 \includegraphics[width=0.955\linewidth]{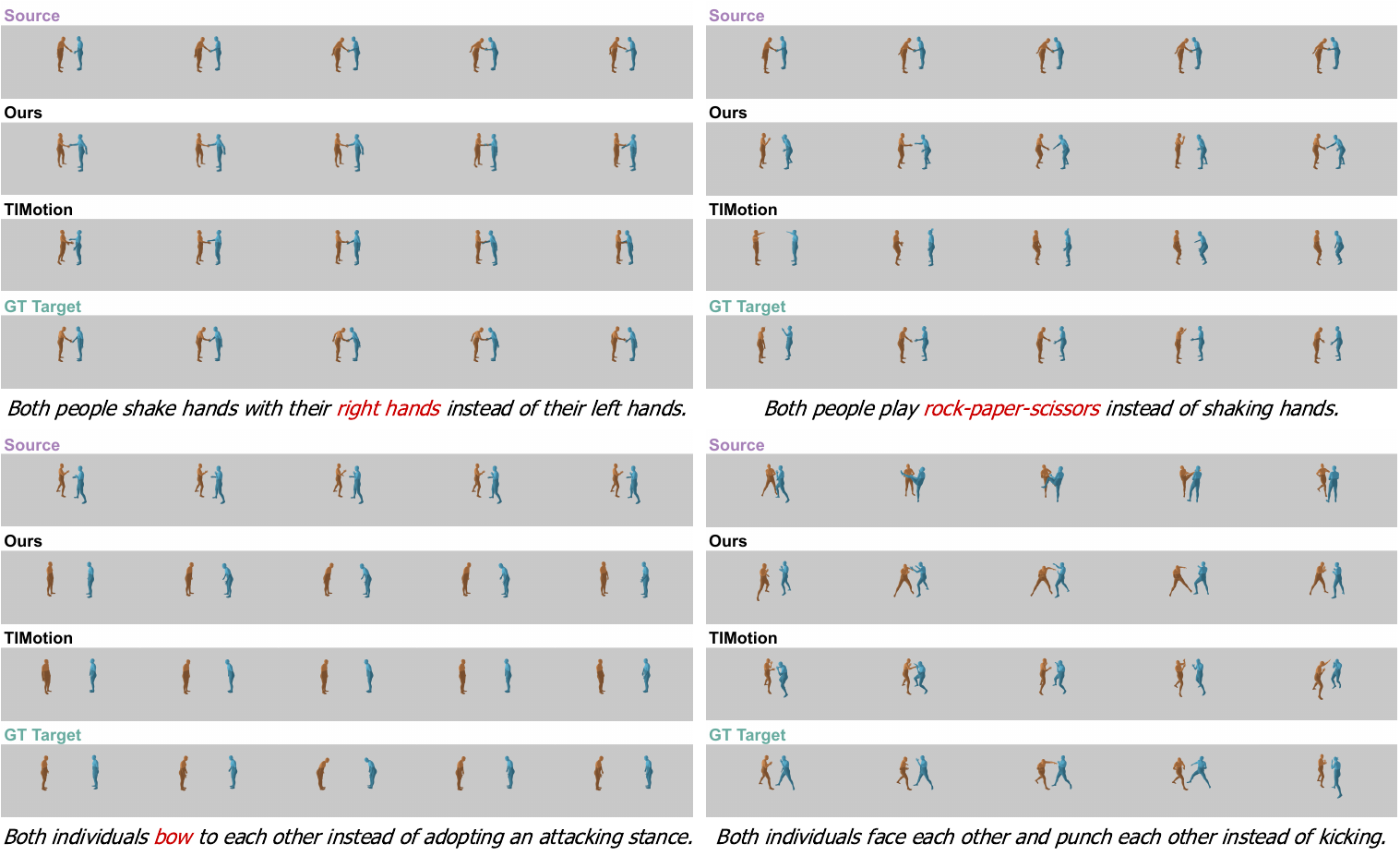}
 \caption{Qualitative results comparison of our \textbf{InterEdit} and TIMotion~\cite{wang2025TIMotion}.}
\label{fig:qualitative_results}
\end{figure*}
\begin{figure*}[h]
 \centering
 \includegraphics[width=\linewidth]{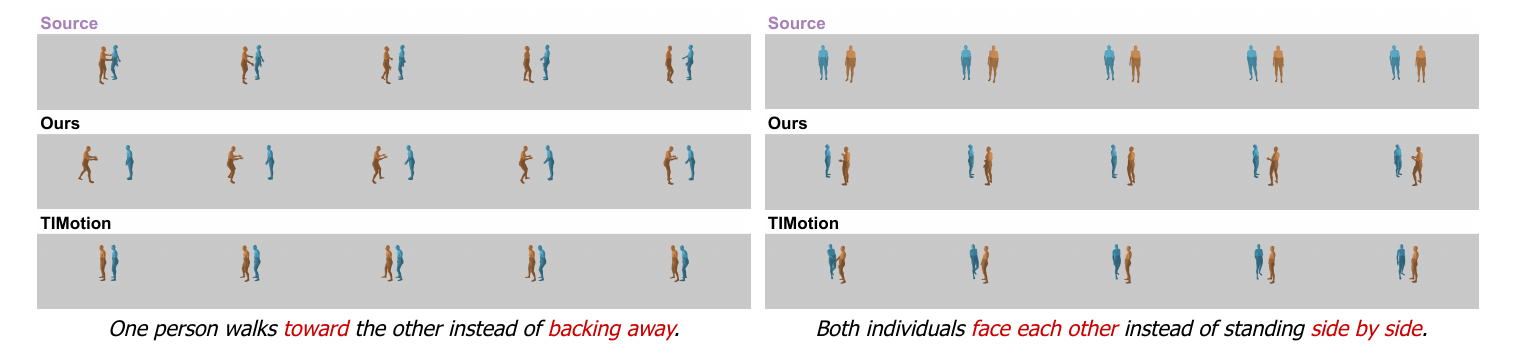}
  \caption{Qualitative results comparison under custom prompts.}
  \label{fig:ood}
\end{figure*}
\subsection{Analysis of the Qualitative Results}
We present qualitative results on test-set prompts (Fig.~\ref{fig:qualitative_results}) and custom prompts (Fig.~\ref{fig:ood}), comparing our results with the TIMotion~\cite{wang2025TIMotion} baseline. The examples cover action-category changes, temporal coordination, and spatial relations.

\noindent\textit{``Rock-paper-scissors instead of shaking hands.''} requires both semantic switch and correct temporal coupling. Our result exhibits synchronized throws, while TIMotion produces offset gestures. This highlights the advantage of regularizing interaction dynamics to enforce synchrony.
For \textit{``Both individuals face each other and punch each other instead of kicking.''}, our result shows an asymmetric attack-defense coupling, matching the target’s intent. TIMotion produces a symmetric response, illustrating the importance of capturing high-level semantics in dual-person editing.
For \textit{``One person walks toward the other instead of backing away.''}, our model accurately reflects the intended approach behavior, while TIMotion fails to capture the directional change, demonstrating the value of semantic-level guidance.

\section{Conclusion}
We introduce the task of \textbf{text-guided multi-human motion editing}, where a model applies instruction-driven changes while preserving source consistency and spatio-temporal coupling. To support this task, we construct a dataset of source-target-edit triplets and establish a benchmark with retrieval-based metrics.
We introduce \textbf{InterEdit3D} dataset and propose \textbf{InterEdit}, a conditional diffusion framework that uses plan tokens for semantic alignment and frequency tokens for interaction-aware DCT descriptors, ensuring precise edits while preserving interaction dynamics.
Our method outperforms single-person editors and interaction generators in instruction adherence, source preservation, and motion realism, providing a foundation for future research in interaction editing and long-horizon motion modifications.

\section*{Acknowledgment}
The project served to prepare the SFB 1574 Circular Factory for the Perpetual Product (project ID: 471687386), approved by the German Research Foundation (DFG, German Research Foundation) with a start date of April 1, 2024. 
This work was also partially supported by the SmartAge project sponsored by the Carl Zeiss Stiftung (P2019-01-003; 2021-2026). 
This work was performed on the HoreKa supercomputer funded by the Ministry of Science, Research and the Arts Baden-Württemberg and by the Federal Ministry of Education and Research. The authors also acknowledge support by the state of Baden-Württemberg through bwHPC and the German Research Foundation (DFG) through grant INST 35/1597-1 FUGG. 
This project was also supported partially by the National Natural Science Foundation of China under Grant No. 62473139, partially by the Hunan Provincial Research and Development Project under Grant No. 2025QK3019, and partially by the Open Research Project of the State Key Laboratory of Industrial Control Technology, China, under Grant No. ICT2025B20. 
\clearpage
\bibliographystyle{splncs04}
\bibliography{main}

\appendix

\section{Impact Statement}
The proposed \textbf{Text-guided Multi-human Motion Editing (TMME)} task and \textbf{InterEdit} framework offer significant contributions to the field of computer vision, particularly in the realm of human motion editing. The ability to edit interactions between multiple humans in 3D motion data, driven by text instructions, opens new possibilities for various applications such as animation, robotics, virtual agents, and training data generation for AI systems. By addressing the challenges associated with editing dual-person motions—especially maintaining spatial and temporal coherence while adhering to semantic instructions—this work paves the way for more intuitive and precise interaction modeling.

The \textbf{InterEdit3D dataset} is the first large-scale collection specifically designed for dual-person motion editing, enabling more accurate evaluation of multi-person interaction models. This dataset's unique pairing of source and target motions with corresponding editing instructions provides a benchmark for future research and promotes progress in the broader community.

The \textbf{InterEdit model}, leveraging advanced techniques like \textit{semantic-aware plan token alignment} and \textit{interaction-oriented frequency token regularization}, achieves state-of-the-art performance in editing accuracy and motion realism. This approach ensures that the fine-grained dynamics of multi-human interactions, such as synchronization and role-switching, are preserved while making instruction-driven changes.

By providing a scalable, practical solution for interaction-aware motion editing, this work has a transformative impact on the development of human-AI collaboration tools, entertainment production, and human-robot interaction systems, all of which require highly detailed, controllable motion editing capabilities. The publicly available code and dataset will facilitate further exploration and innovation, broadening the impact of this research across industries that rely on realistic, editable motion data.

\begin{figure*}[!t]
  \centering

  \begin{subfigure}[t]{0.40\textwidth}
    \centering
    \includegraphics[width=\linewidth]{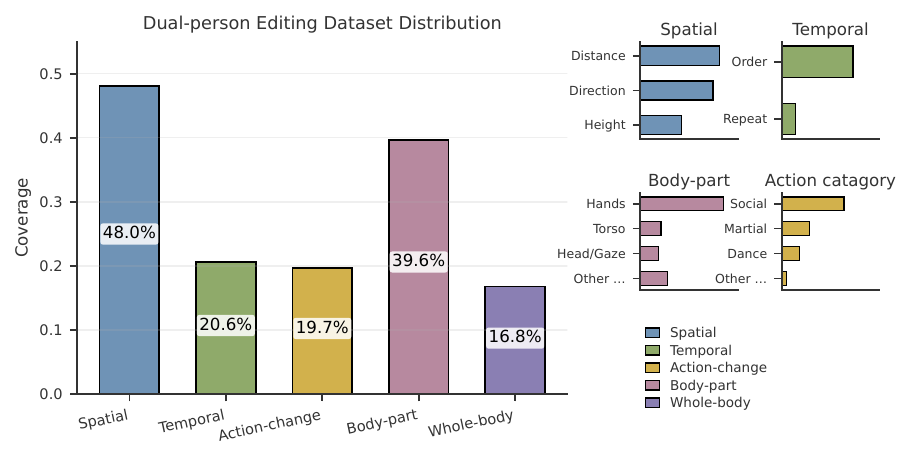}
    \caption{Coverage of data types}
    \label{fig:dist_a}
  \end{subfigure}\hfill
  \begin{subfigure}[t]{0.58\textwidth}
    \centering
    \includegraphics[width=\linewidth]{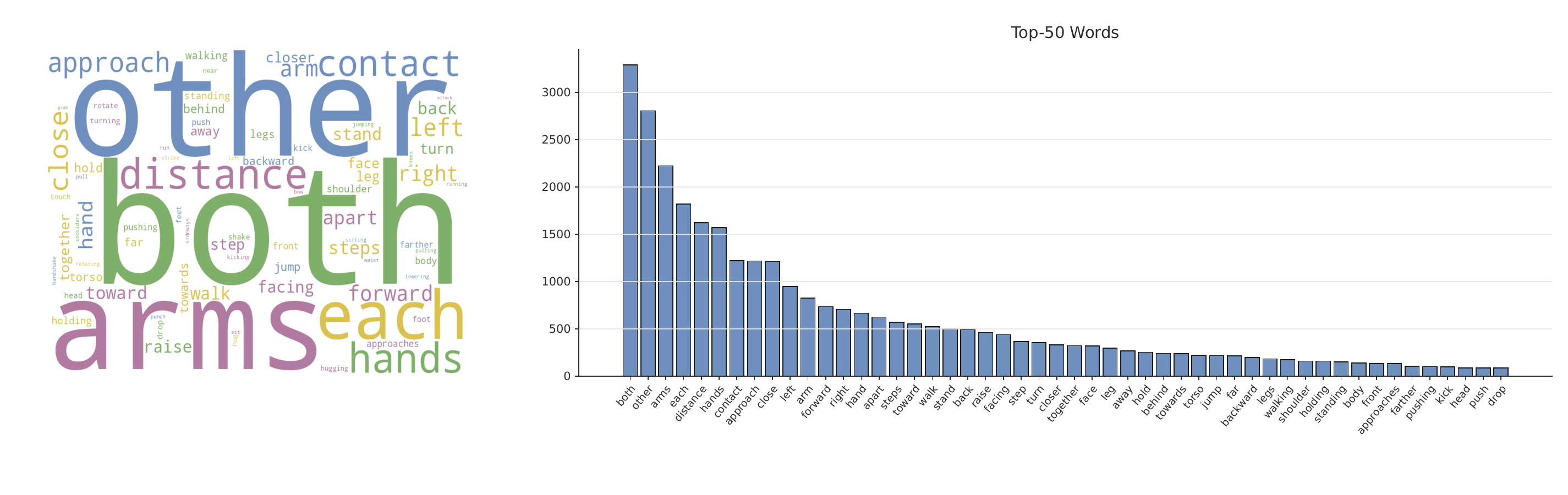}
    \caption{Word cloud of dataset and Top-50 word distribution}
    \label{fig:dist_b}
  \end{subfigure}
\vskip-2ex
  \caption{Dataset statistics of InterEdit3D.
(a) Coverage of semantic dimensions (Spatial, Temporal, Action-change, Body-part, Whole-body), showing dominance of spatial and temporal edits.
(b) Word cloud and Top-50 distribution, highlighting frequent interaction-related and spatial terms.}
\vskip-2ex
  \label{fig:dataset_stats}
\end{figure*}

\section{More Details of the Proposed InterEdit3D Dataset}

\noindent\textbf{Statistics of the InterEdit3D.} We deliver the statistics of the \textit{InterEdit3D} dataset in Fig.~\ref{fig:dataset_stats}, including semantic coverage analysis, instruction distribution, and lexical trend.
\underline{Semantic Coverage Analysis:}
To quantify the annotation diversity of InterEdit3D, we conduct a rule-based analysis over all edit texts and measure coverage along four interaction-relevant dimensions: \emph{Spatial relations} (direction, distance, relative position, height), \emph{Temporal structure} (order, phases, repetition), \emph{Action-category change} (\eg, social, martial, dance), and \emph{Body-part constraints}. These dimensions are not mutually exclusive; thus, we report coverage rates rather than a partition summing to 100\%.
\underline{Instruction Distribution:} The dataset is dominated by spatial and temporal cues, while still containing substantial body-part and action-change edits. Spatial instructions frequently describe distance variation (approach/move apart) and orientation changes (face-to-face, left/right, front/behind). Temporal edits mainly specify ordering (before/after/start/end) and repetition (again/twice). Action-category changes reflect both fine-grained relational edits and larger semantic shifts in interaction style. Body-part constraints commonly emphasize hands and arms, consistent with contact-rich interactions (\eg, holding, passing, pushing). We additionally define \emph{whole-body} instructions as those without explicit body-part keywords but containing spatial, temporal, or action-change cues; a notable portion of annotations fall into this category, indicating global interaction-level edits.
\underline{Lexical Trends:} Word-frequency analysis further confirms strong interaction semantics, with frequent spatial terms (left/right/back/forward), relational cues (contact/close/apart), and dual-person references (both/other), alongside diverse action verbs spanning social, martial, and dance domains.

\section{More Samples from InterEdit3D}
The \textbf{InterEdit3D} dataset is designed to facilitate the development of models for dual-person motion editing with text-guided instructions. It consists of \textit{source-target-text triplets}, where each triplet pairs a source motion with a target motion and an associated editing instruction. This section provides further details on representative samples from the dataset, showcasing the range of interactions and the challenges involved in editing multi-human motion sequences.

\begin{figure*}[h]
  \centering
  \setlength{\tabcolsep}{0pt} 
  \renewcommand{\arraystretch}{1.0}

\begin{tabular}{@{}p{0.495\textwidth}@{\hspace{2pt}}p{0.495\textwidth}@{}}
\begin{minipage}[t]{\linewidth}
  \centering
  \begin{overpic}[width=\linewidth]{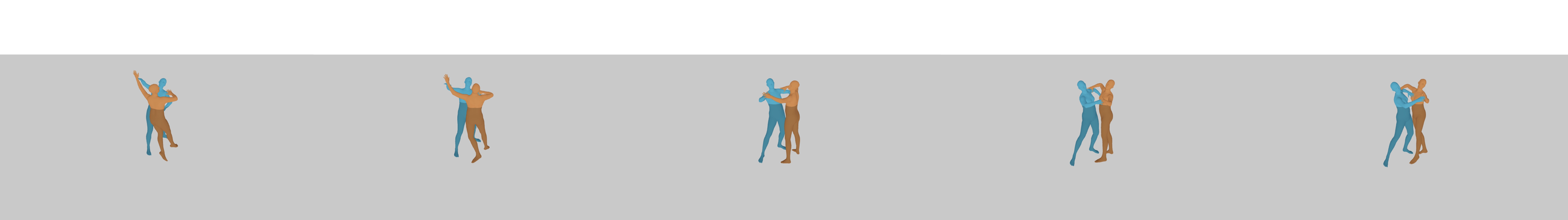}
    \put(1,11){{\scriptsize\bfseries Source}}
  \end{overpic}\\[1mm]
  \begin{overpic}[width=\linewidth]{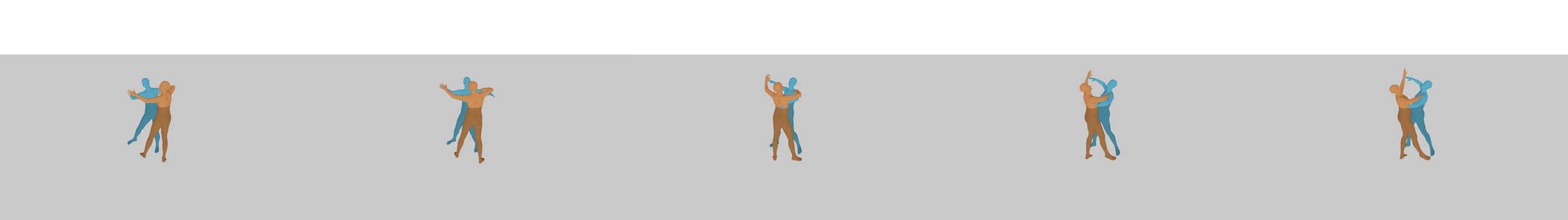}
    \put(1,11){{\scriptsize\bfseries Target}}
  \end{overpic}\\[1mm]
  \medskip
  \parbox{0.98\linewidth}{\centering\small
    \textit{Both individuals hold each other in a ballroom dance position and \textcolor{red!70!black}{rotate clockwise} in place instead of \textcolor{red!70!black}{rotating counterclockwise}.}
  }
\end{minipage}
&
\begin{minipage}[t]{\linewidth}
  \centering
  \begin{overpic}[width=\linewidth]{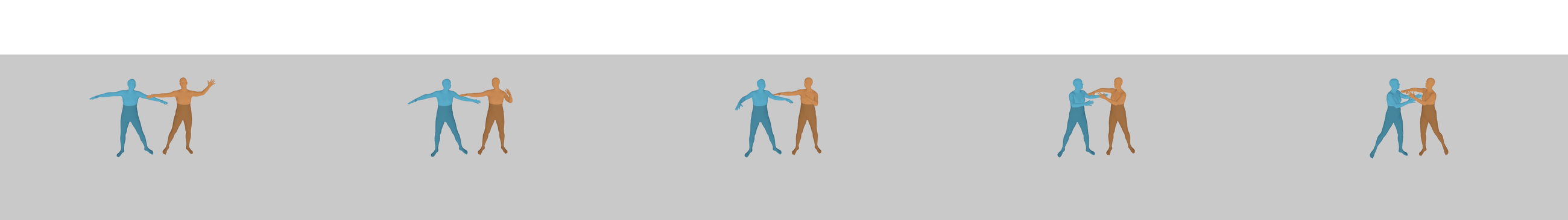}
    \put(1,11){{\scriptsize\bfseries Source}}
  \end{overpic}\\[1mm]
  \begin{overpic}[width=\linewidth]{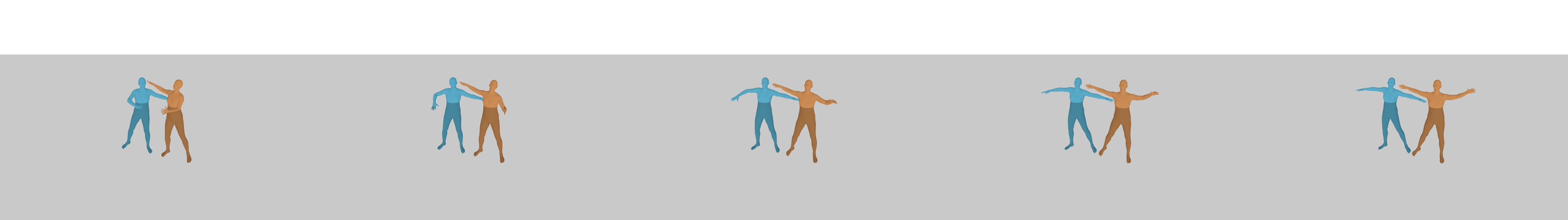}
    \put(1,11){{\scriptsize\bfseries Target}}
  \end{overpic}\\[1mm]
  \medskip
  \parbox{0.98\linewidth}{\centering\small
    \textit{The two individuals stand side by side and extend their outer arms \textcolor{red!70!black}{from inward to outward}, rather than \textcolor{red!70!black}{the opposite direction}.}
  }
\end{minipage}
\\
\noalign{\vskip 4mm}

\begin{minipage}[t]{\linewidth}
  \centering
  \begin{overpic}[width=\linewidth]{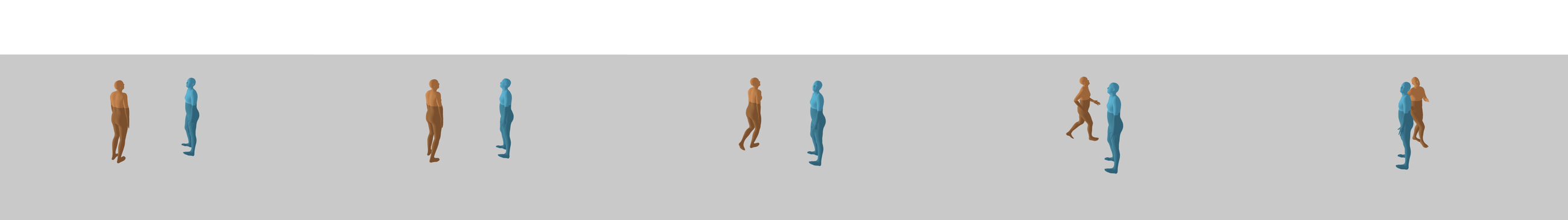}
    \put(1,11){{\scriptsize\bfseries Source}}
  \end{overpic}\\[1mm]
  \begin{overpic}[width=\linewidth]{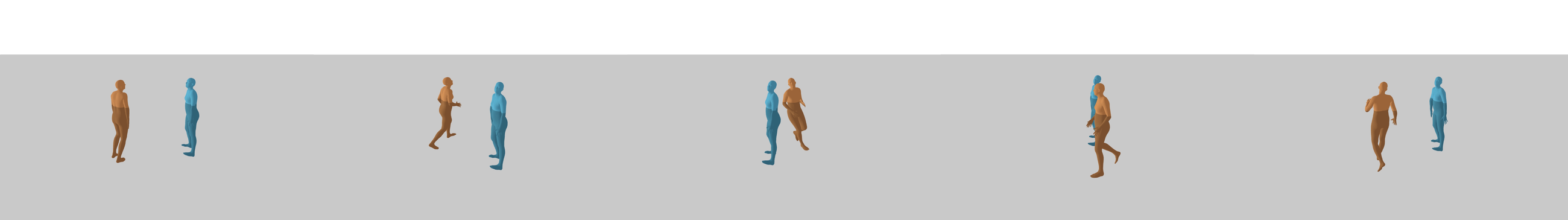}
    \put(1,11){{\scriptsize\bfseries Target}}
  \end{overpic}\\[1mm]
  \medskip
  \parbox{0.98\linewidth}{\centering\small
    \textit{Person 2 should \textcolor{red!70!black}{continue running around} Person 1, completing \textcolor{red!70!black}{a full circle} around Person 1.}
  }
\end{minipage}
&
\begin{minipage}[t]{\linewidth}
  \centering
  \begin{overpic}[width=\linewidth]{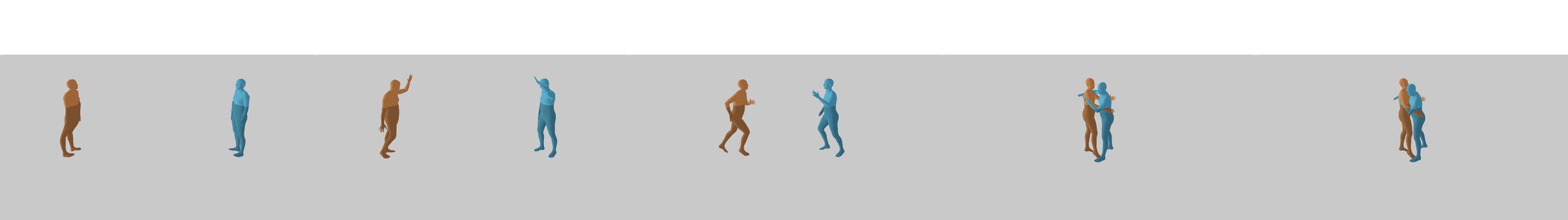}
    \put(1,11){{\scriptsize\bfseries Source}}
  \end{overpic}\\[1mm]
  \begin{overpic}[width=\linewidth]{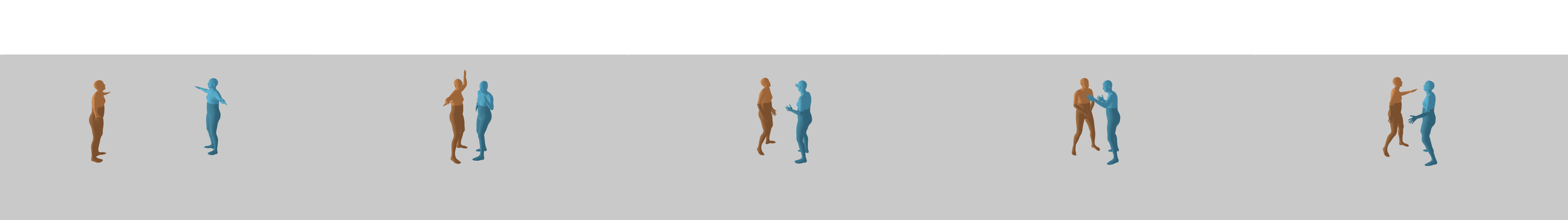}
    \put(1,11){{\scriptsize\bfseries Target}}
  \end{overpic}\\[1mm]
  \medskip
  \parbox{0.98\linewidth}{\centering\small
    \textit{The two individuals approach each other face to face and \textcolor{red!70!black}{engage in a heated argument}, rather than \textcolor{red!70!black}{hugging}.}
  }
\end{minipage}
\\
\noalign{\vskip 4mm}

\begin{minipage}[t]{\linewidth}
  \centering
  \begin{overpic}[width=\linewidth]{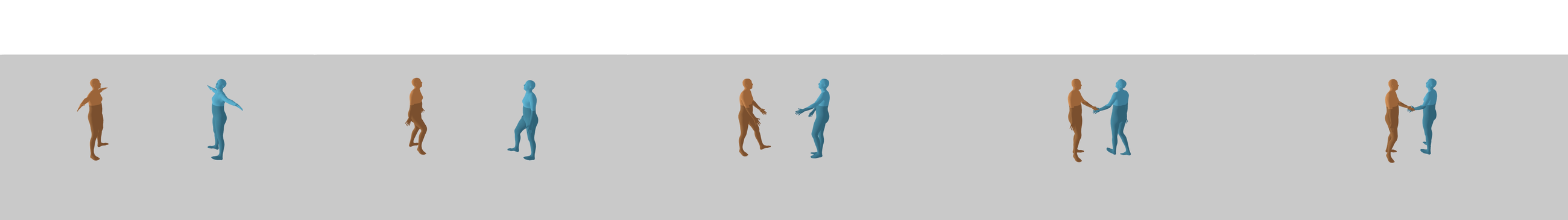}
    \put(1,11){{\scriptsize\bfseries Source}}
  \end{overpic}\\[1mm]
  \begin{overpic}[width=\linewidth]{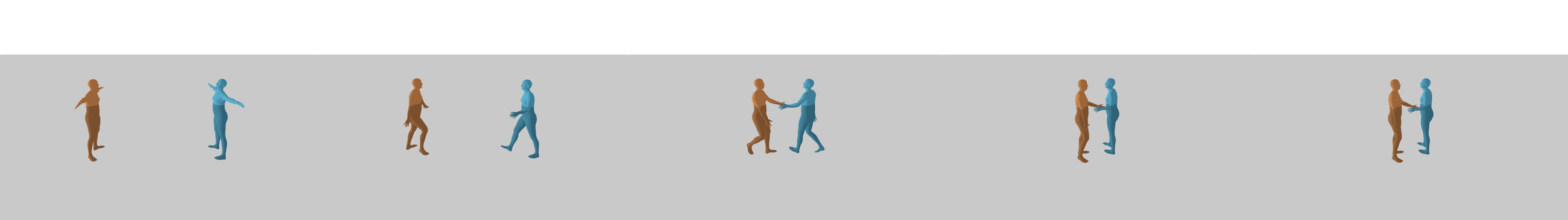}
    \put(1,11){{\scriptsize\bfseries Target}}
  \end{overpic}\\[1mm]
  \medskip
  \parbox{0.98\linewidth}{\centering\small
    \textit{Both individuals maintain the handshake for a \textcolor{red!70!black}{longer duration}.}
  }
\end{minipage}
&
\begin{minipage}[t]{\linewidth}
  \centering
  \begin{overpic}[width=\linewidth]{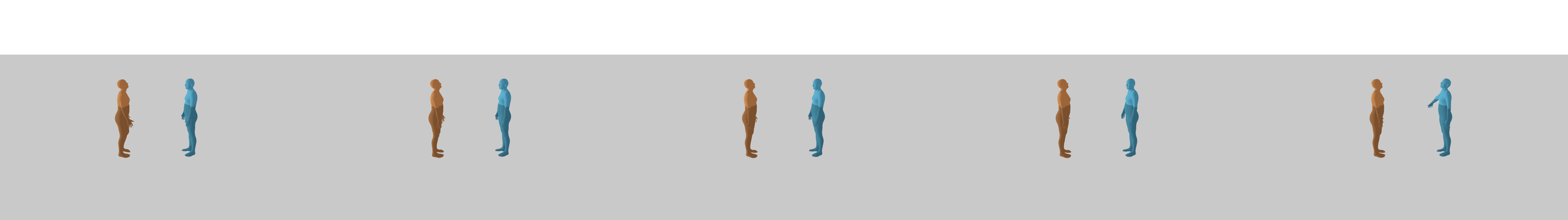}
    \put(1,11){{\scriptsize\bfseries Source}}
  \end{overpic}\\[1mm]
  \begin{overpic}[width=\linewidth]{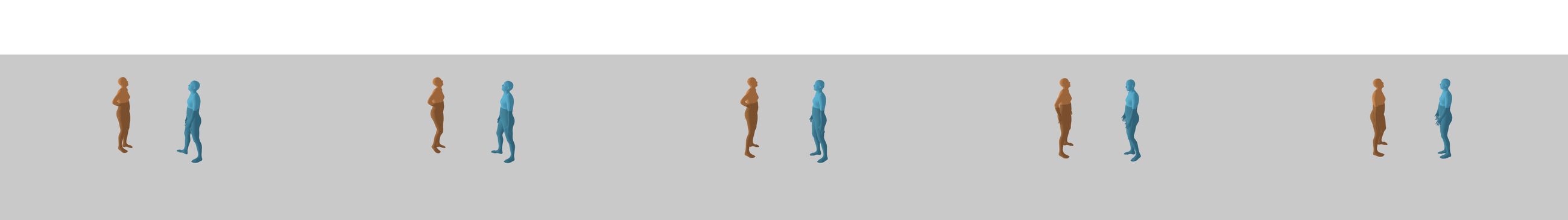}
    \put(1,11){{\scriptsize\bfseries Target}}
  \end{overpic}\\[1mm]
  \medskip
  \parbox{0.98\linewidth}{\centering\small
    \textit{Keep the current pose while one person \textcolor{red!70!black}{lowers the raised arm}.}
  }
\end{minipage}
\\

  \end{tabular}

  \caption{Dataset samples: We display source motions with target motions from our \textit{InterEdit3D} dataset, together with their corresponding
text annotations}
  \label{fig:dataset_samples}
\end{figure*}

Each sample involves a dual-person interaction with specific textual instructions aimed at modifying the motion while preserving the overall coordination and realism of the interaction. Below are examples from the dataset that demonstrate the diversity of actions and interactions included:

The \textbf{InterEdit3D} dataset includes a variety of motion editing tasks with corresponding text annotations. These samples from the dataset are displayed in Fig.~\ref{fig:dataset_samples}, where we show the source motions alongside the target motions generated from the given text instructions. Each pair demonstrates how a given instruction modifies the interaction dynamics while preserving the overall temporal and spatial relationships.

\noindent\textbf{1. Ballroom Dance Position with Clockwise Rotation.}
In the first example, two individuals are initially holding each other in a ballroom dance position and rotating counterclockwise. The editing instruction asks them to rotate clockwise instead. This modification requires adjusting the relative positioning and rotation direction of both individuals while maintaining the smooth coordination of their movements.

\noindent\textbf{2. Standing Side by Side with Arm Extension.}
In this example, two individuals are instructed to stand side by side and extend their outer arms from inward to outward, rather than the opposite direction. This change involves modifying the arm positions while preserving the spatial and temporal coherence between the two individuals.

\noindent\textbf{3. Running Around Person 1.}
The third example features a person running around another individual. The instruction asks Person 2 to continue running around Person 1, completing a full circle. This modification requires precise adjustments in the trajectory of Person 2 while maintaining the spatial relationship between the two participants.

\noindent\textbf{4. Engaging in a Heated Argument Instead of Hugging.}
In this case, the source motion depicts two individuals hugging. The editing instruction changes the interaction to a heated argument, with both individuals approaching each other face-to-face. This change requires careful adjustment of both body positioning and gesture, ensuring that the motion dynamics align with the new interaction type.

\noindent\textbf{5. Maintaining the Handshake for a Longer Duration.}
Here, two individuals are shown shaking hands. The instruction asks them to maintain the handshake for a longer duration. This sample demonstrates how the editing process focuses on extending the interaction time while preserving the motion dynamics of the handshake.

\noindent\textbf{6. Lowering the Raised Arm While Maintaining the Current Pose.}
In the final sample, the source motion features two individuals with one of them raising their arm. The instruction requires the individual to keep the current pose while lowering the raised arm. This adjustment focuses on modifying a specific aspect of the motion while preserving the overall body posture and interaction.
These samples from the \textit{InterEdit3D} dataset highlight the variety of interaction edits that can be applied with text instructions. The dataset includes a range of tasks from simple positional changes to more complex alterations involving dynamic interaction shifts, such as role changes and coordinated actions.

\begin{table}[h]
\centering
\resizebox{\linewidth}{!}{
    \begin{tabular}{l c ccc ccc}
    \toprule
    & & \multicolumn{3}{c}{generated-to-source retrieval (\%) $\uparrow$} & \multicolumn{3}{c}{ \textbf{generated-to-target retrieval (\%) $\uparrow$}} \\
    \cmidrule(lr){3-5} \cmidrule(lr){6-8}
    Method & FID $\downarrow$ & R@1 & R@2 & R@3 & \textbf{R@1} & \textbf{R@2} & \textbf{R@3} \\
    \midrule
    plan layer=2 &
    0.3718$\pm$0.0028 &
    17.58$\pm$0.33 & 24.05$\pm$0.44 & 28.31$\pm$0.38 &
    \cellcolor{gray!15}30.26$\pm$0.49 & \cellcolor{gray!15}40.32$\pm$0.40 & \cellcolor{gray!15}46.51$\pm$0.36 \\
    plan layer=3 &
    0.3707$\pm$0.0029 &
    17.08$\pm$0.41 & 24.04$\pm$0.46 & 29.32$\pm$0.49 &
    \cellcolor{gray!15}\textbf{30.82$\pm$0.43} & \cellcolor{gray!15}\textbf{40.84$\pm$0.70} & \cellcolor{gray!15}\textbf{47.65$\pm$0.59} \\
    plan layer=4 &
    0.3369$\pm$0.0022 &
    14.70$\pm$0.41 & 21.44$\pm$0.35 & 26.09$\pm$0.39 &
    \cellcolor{gray!15}30.53$\pm$0.43 & \cellcolor{gray!15}39.84$\pm$0.40 & \cellcolor{gray!15}45.57$\pm$0.44 \\
    plan layer=5 &
    0.3719$\pm$0.0027 &
    16.77$\pm$0.32 & 23.16$\pm$0.39 & 27.45$\pm$0.48 &
    \cellcolor{gray!15}29.65$\pm$0.35 & \cellcolor{gray!15}40.17$\pm$0.48 & \cellcolor{gray!15}46.57$\pm$0.60 \\
    \bottomrule
    \end{tabular}
}
\caption{Ablation study regarding the location to conduct semantic-aware plan token alignment (freq layer=5, drop=0.04, mean, 95\% CI).}
\label{tab:exp_plan_layer}
\end{table}

\begin{table}[h]
\centering
\resizebox{\linewidth}{!}{
    \begin{tabular}{l c ccc ccc}
    \toprule
    & & \multicolumn{3}{c}{generated-to-source retrieval (\%) $\uparrow$} & \multicolumn{3}{c}{ \textbf{generated-to-target retrieval (\%) $\uparrow$}} \\
    \cmidrule(lr){3-5} \cmidrule(lr){6-8}
    Loss type & FID $\downarrow$ & R@1 & R@2 & R@3 & \textbf{R@1} & \textbf{R@2} & \textbf{R@3} \\
    \midrule
    InfoNCE &
    0.3707$\pm$0.0029 &
    17.08$\pm$0.41 & 24.04$\pm$0.46 & 29.32$\pm$0.49 &
    \cellcolor{gray!15}\textbf{30.82$\pm$0.43} & \cellcolor{gray!15}\textbf{40.84$\pm$0.70} & \cellcolor{gray!15}\textbf{47.65$\pm$0.59} \\
    Cosine &
    0.3410$\pm$0.0032 &
    15.51$\pm$0.30 & 22.54$\pm$0.37 & 27.10$\pm$0.43 &
    \cellcolor{gray!15}30.03$\pm$0.54 & \cellcolor{gray!15}39.10$\pm$0.42 & \cellcolor{gray!15}44.88$\pm$0.44 \\
    MSE &
    0.3532$\pm$0.0033 &
    14.89$\pm$0.28 & 22.54$\pm$0.29 & 26.78$\pm$0.40 &
    \cellcolor{gray!15}28.07$\pm$0.54 & \cellcolor{gray!15}37.80$\pm$0.46 & \cellcolor{gray!15}45.13$\pm$0.55 \\
    \bottomrule
    \end{tabular}
}
\caption{Ablation study regarding the selection of loss for semantic-aware plan token alignment (mean, 95\% CI).}
\label{tab:exp_plan_loss_type}
\end{table}

\section{More Analysis of the Ablation Study}
\noindent\textbf{Where to apply plan-token supervision.} Table~\ref{tab:exp_plan_layer} varies the transformer block where the plan loss is applied. We find that supervising plan tokens at an intermediate block leads to the strongest g2t/g2s retrieval and competitive FID. This is consistent with the role of plan tokens as high-level semantic carriers: at early layers, representations are still too low-level to reliably encode editing intent, whereas at very late layers the denoising trajectory has already committed to fine details and semantic guidance becomes less effective. Intermediate-layer supervision strikes a favorable balance, enabling the latent plan representation to steer generation while still leaving sufficient capacity for downstream refinement.

\noindent\textbf{Plan loss type.} Table~\ref{tab:exp_plan_loss_type} compares different plan-loss formulations. InfoNCE yields the best overall retrieval and FID trade-off compared to cosine and MSE losses. This result supports using a contrastive objective for semantic alignment: it better preserves discriminative structure in the latent space and encourages the learned plan representation to match target semantics in a way that generalizes across diverse interactions and edit instructions.

\noindent\textbf{Sensitivity to loss weights.} 
Finally, Table~\ref{tab:exp_freq_loss_weight_001_wp} and Table~\ref{tab:exp_plan_loss_weight_003_wf} analyze the sensitivity to the weighting of the plan loss and frequency loss. Performance is robust within a reasonable range, with the best results achieved near our default settings ($\lambda_{\text{f}}$=0.01, $\lambda_{\text{p}}$=0.03). 
Over-weighting either loss can slightly degrade FID or retrieval, suggesting that both semantic alignment and frequency alignment should act as auxiliary regularizers rather than dominating the diffusion objective.

\begin{table}[h]
\centering
\resizebox{\linewidth}{!}{
    \begin{tabular}{l c ccc ccc}
    \toprule
    & & \multicolumn{3}{c}{generated-to-source retrieval (\%) $\uparrow$} & \multicolumn{3}{c}{ \textbf{generated-to-target retrieval (\%) $\uparrow$}} \\
    \cmidrule(lr){3-5} \cmidrule(lr){6-8}
    Loss weight & FID $\downarrow$ & R@1 & R@2 & R@3 & \textbf{R@1} & \textbf{R@2} & \textbf{R@3} \\
    \midrule
     $\lambda_p$=0.03 &
    0.3707$\pm$0.0029 &
    17.08$\pm$0.41 & 24.04$\pm$0.46 & 29.32$\pm$0.49 &
    \cellcolor{gray!15}30.82$\pm$0.43 & \cellcolor{gray!15}40.84$\pm$0.70 & \cellcolor{gray!15}47.65$\pm$0.59 \\
    $\lambda_p$=0.3 &
    0.3585$\pm$0.0028 &
    16.04$\pm$0.30 & 22.69$\pm$0.33 & 26.86$\pm$0.32 &
    \cellcolor{gray!15}31.21$\pm$0.42 & \cellcolor{gray!15}41.41$\pm$0.32 & \cellcolor{gray!15}47.39$\pm$0.37 \\
     $\lambda_p$=3 &
    0.3620$\pm$0.0029 &
    16.94$\pm$0.38 & 24.50$\pm$0.42 & 30.46$\pm$0.40 &
    \cellcolor{gray!15}29.72$\pm$0.49 & \cellcolor{gray!15}39.65$\pm$0.55 & \cellcolor{gray!15}46.50$\pm$0.56 \\
    \bottomrule
    \end{tabular}
}
\caption{Ablation study of $\lambda_p$ ($\lambda_f$= 0.01, mean, 95\% CI).}
\label{tab:exp_freq_loss_weight_001_wp}
\end{table}

\begin{table}[h]
\centering
\resizebox{\linewidth}{!}{
    \begin{tabular}{l c ccc ccc}
    \toprule
    & & \multicolumn{3}{c}{generated-to-source retrieval (\%) $\uparrow$} & \multicolumn{3}{c}{ \textbf{generated-to-target retrieval (\%) $\uparrow$}} \\
    \cmidrule(lr){3-5} \cmidrule(lr){6-8}
    Loss weight & FID $\downarrow$ & R@1 & R@2 & R@3 & \textbf{R@1} & \textbf{R@2} & \textbf{R@3} \\
    \midrule
    $\lambda_f$=0.01 &
    0.3707$\pm$0.0029 &
    17.08$\pm$0.41 & 24.04$\pm$0.46 & 29.32$\pm$0.49 &
    \cellcolor{gray!15}30.82$\pm$0.43 & \cellcolor{gray!15}40.84$\pm$0.70 & \cellcolor{gray!15}47.65$\pm$0.59 \\
    $\lambda_f$=0.1 &
    0.3560$\pm$0.0029 &
    15.69$\pm$0.18 & 22.10$\pm$0.35 & 27.29$\pm$0.44 &
    \cellcolor{gray!15}29.62$\pm$0.61 & \cellcolor{gray!15}40.03$\pm$0.51 & \cellcolor{gray!15}46.07$\pm$0.55 \\
    $\lambda_f$=1 &
    0.3492$\pm$0.0030 &
    16.08$\pm$0.38 & 22.30$\pm$0.39 & 26.50$\pm$0.50 &
    \cellcolor{gray!15}30.53$\pm$0.27 & \cellcolor{gray!15}40.89$\pm$0.53 & \cellcolor{gray!15}47.54$\pm$0.57 \\
    \bottomrule
    \end{tabular}
}
\caption{Ablation study of the $\lambda_f$ ($\lambda_p$=0.03, mean, 95\% CI).}
\label{tab:exp_plan_loss_weight_003_wf}
\end{table}

\noindent\textbf{Two-branch vs. three-branch SCFG.}
We also ablate two-branch SCFG against three-branch SCFG.
The two-branch SCFG consists of a joint-conditioned branch and an unconditional branch, while the three-branch SCFG additionally introduces a source-only branch to separate source guidance from text guidance.
Accordingly, the two-branch formulation requires two denoising predictions per diffusion step, whereas the three-branch formulation requires three.

From Table~\ref{tab:CFG}, we observe that the two designs are highly comparable overall, with no stable and substantial advantage for the three-branch variant.
While adding the source-only branch can slightly improve a subset of metrics, the improvements remain limited and are not consistently reflected across all evaluation criteria.
This indicates that the extra branch does not lead to sufficiently significant gains in our setting.

Based on this observation, we choose the two-branch SCFG as the default setting in the final model.
Compared with three-branch SCFG, it has a simpler guidance design and lower inference cost, since only two denoising predictions are required during sampling.
Considering that its performance remains competitive while the sampling overhead is lower, two-branch SCFG provides a more favorable trade-off between simplicity, efficiency, and editing quality.

\begin{table}[h]
\centering
\resizebox{\linewidth}{!}{
    \begin{tabular}{l c ccc ccc}
    \toprule
    & & \multicolumn{3}{c}{generated-to-source retrieval (\%) $\uparrow$} & \multicolumn{3}{c}{ \textbf{generated-to-target retrieval (\%) $\uparrow$}} \\
    \cmidrule(lr){3-5} \cmidrule(lr){6-8}
    Variants & FID $\downarrow$ & R@1 & R@2 & R@3 & \textbf{R@1} & \textbf{R@2} & \textbf{R@3} \\
    \midrule
    two-branch &
    0.3707$\pm$0.0029 &
    17.08$\pm$0.41 & 24.04$\pm$0.46 & 29.32$\pm$0.49 &
    \cellcolor{gray!15}30.82$\pm$0.43 & \cellcolor{gray!15}40.84$\pm$0.70 & \cellcolor{gray!15}47.65$\pm$0.59 \\
    three-branch &
    0.3713$\pm$0.0043 &
    17.27$\pm$0.34 & 24.15$\pm$0.43 & 28.82$\pm$0.37 &
    \cellcolor{gray!15}31.26$\pm$0.49 & \cellcolor{gray!15}41.19$\pm$0.41 & \cellcolor{gray!15}47.42$\pm$0.36 \\
    \bottomrule
    \end{tabular}
}
\caption{Ablation study of two-branch \textit{vs.} three-branch SCFG (mean, $95\%$ CI).}
\label{tab:CFG}
\end{table}

\noindent\textbf{Target-side auxiliary supervision.}
Our plan and frequency alignment losses use target-side signals only as auxiliary training targets: the plan tokens are aligned to the frozen teacher embedding of the ground-truth target motion, while the frequency tokens are aligned to target DCT band-energy descriptors. These targets define the desired edited interaction in semantic and temporal-frequency spaces, but they are never provided as input conditions to the denoising model. At inference, both plan and frequency tokens are inferred from the source motion, the editing instruction, and the denoising process. Therefore, target-side supervision serves as training-time guidance for learning internal editing representations rather than test-time privileged information.

To further verify this formulation, we weaken the target-derived auxiliary signals by adding Gaussian noise with standard deviation $0.3$ during training. As shown in Table~\ref{tab:weak_target_supervision}, weakening these signals degrades both generated-to-target and generated-to-source retrieval. Compared with the default setting, the average g2t and g2s scores over R@1/R@2/R@3 drop relatively by $3.1\%$ and $4.8\%$, respectively. This supports the role of target teacher embeddings and target frequency descriptors as meaningful auxiliary supervision for learning semantic editing intent and interaction dynamics, rather than as privileged inference-time information.

\begin{table}[h]
\centering
\small
\setlength{\tabcolsep}{6pt}
\renewcommand{\arraystretch}{1.08}
\begin{tabular}{l ccc ccc}
\toprule
& \multicolumn{3}{c}{\shortstack{generated-to-source (\%) $\uparrow$}}
& \multicolumn{3}{c}{\shortstack{\textbf{generated-to-target} (\%) $\uparrow$}} \\
\cmidrule(lr){2-4} \cmidrule(lr){5-7}
Setting
& R@1 & R@2 & R@3
& \textbf{R@1} & \textbf{R@2} & \textbf{R@3} \\
\midrule
Default
& 17.08 & 24.04 & 29.32
& \cellcolor{gray!15}\textbf{30.82}
& \cellcolor{gray!15}\textbf{40.84}
& \cellcolor{gray!15}\textbf{47.65} \\
Noisy target sup.
& 16.12 & 23.25 & 27.71
& \cellcolor{gray!15}28.76
& \cellcolor{gray!15}40.37
& \cellcolor{gray!15}46.45 \\
\bottomrule
\end{tabular}
\caption{Robustness analysis of target-side auxiliary supervision. ``Noisy target sup.'' denotes weakening target-derived supervision with Gaussian noise of standard deviation $0.3$. Target-derived signals are used only for auxiliary training losses and are never provided as model input at inference.}
\label{tab:weak_target_supervision}
\end{table}

\section{Additional Experiments and Discussions}
\label{sec:additional_experiments}

\subsection{Metric Discussion}
We use generated-to-target retrieval (g2t), generated-to-source retrieval (g2s), and FID as the main quantitative metrics. The g2t metric measures whether the generated motion matches the target interaction semantics implied by the editing instruction, while g2s measures whether the edited result remains anchored to the source motion. FID evaluates motion realism in the learned motion embedding space. These metrics enable scalable comparison across models, but they remain proxy measures and may not fully capture perceptual interaction quality, fine-grained contact, or subtle gesture semantics. We therefore further provide human evaluation in Sec.~\ref{sec:human_eval}.

\subsection{Generalization Beyond the Editing Benchmark}
To examine whether our token-level alignment design generalizes beyond InterEdit3D, we additionally evaluate it on the standard InterHuman two-person generation benchmark. In this setting, the model generates two-person interactions from text without source-motion conditioning. As shown in Table~\ref{tab:interhuman_generation_supp}, our model improves R-Precision over InterGen and TIMotion variants, suggesting that semantic-aware plan tokens and interaction-aware frequency tokens are also beneficial for general two-person motion modeling.

\begin{table}[h]
\centering
\resizebox{0.9\linewidth}{!}{
\begin{tabular}{lcccc}
\toprule
R-Precision & InterGen & TIMotion+Transformer & TIMotion+RWKV & \textbf{Ours} \\
\midrule
R@1$\uparrow$ & 0.371$\pm$0.010 & 0.491$\pm$0.005 & 0.501$\pm$0.005 & \textbf{0.523$\pm$0.004} \\
R@2$\uparrow$ & 0.515$\pm$0.012 & 0.648$\pm$0.004 & 0.656$\pm$0.006 & \textbf{0.665$\pm$0.005} \\
R@3$\uparrow$ & 0.624$\pm$0.010 & 0.724$\pm$0.004 & 0.734$\pm$0.006 & \textbf{0.753$\pm$0.004} \\
\bottomrule
\end{tabular}
}
\caption{Generalization evaluation on the InterHuman two-person generation benchmark.}
\label{tab:interhuman_generation_supp}
\end{table}

\section{Details Regarding the Dataset Creation and Annotation}

\begin{figure}[!t]
    \centering
    \includegraphics[width=\linewidth]{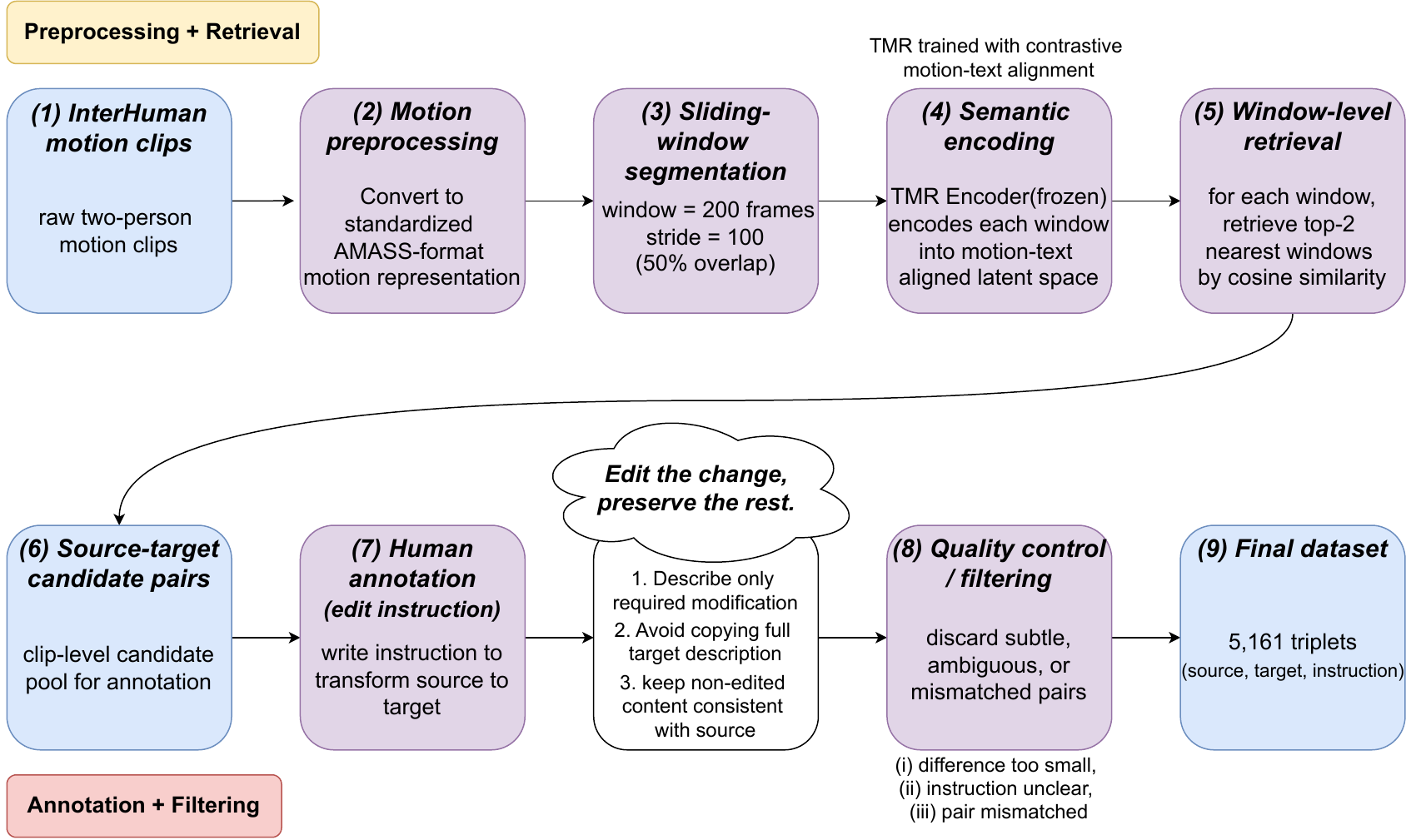}
    \caption{Data collection and annotation pipeline for building our dual-person motion editing dataset.}
    \label{fig:data_pipeline}
\end{figure}

This appendix details our data collection and annotation pipeline, summarized in Fig.~\ref{fig:data_pipeline}.
Starting from raw two-person interaction motions, we mine source--target candidates via window-level retrieval in a motion--text aligned embedding space, and then curate high-quality editing triplets through human instruction writing and quality control.

\paragraph{Motion preprocessing.}
Given a single-person motion of dual-person motions, we convert the raw representation into an AMASS-style feature format that is compatible with the pretrained TMR motion encoder~\cite{petrovich2023tmrtexttomotionretrievalusing} (Fig.~\ref{fig:data_pipeline}-(2)).
This step extracts standardized motion features (\textit{e.g.}, root translation/orientation and pose parameters) and applies dataset statistics for normalization, ensuring consistent scale and distribution across clips.

\paragraph{Sliding-window segmentation.}
To support retrieval for variable-length clips and increase matching robustness, we segment each motion clip into overlapping temporal windows (Fig.~\ref{fig:data_pipeline}-(3)).
Each window serves as an individual retrieval unit, allowing semantically similar interaction segments to be matched even when full clips differ in duration or contain multiple phases.

\paragraph{Semantic motion encoding.}
We embed each window into a motion--text aligned latent space using a frozen TMR motion encoder~\cite{petrovich2023tmrtexttomotionretrievalusing} (Fig.~\ref{fig:data_pipeline}-(4)).
Since TMR is trained with contrastive motion--text alignment, its embeddings are semantically meaningful and thus well-suited for nearest-neighbor retrieval.

\paragraph{Window-level retrieval and candidate pool construction.}
We perform retrieval at the window level.
For each query window, we compute cosine similarity to all database windows and retrieve its top-$2$ nearest neighbors in the embedding space (Fig.~\ref{fig:data_pipeline}-(5)).
Crucially, for every retrieved match we record both the clip index and the window identifier of the query and the neighbor, \textit{i.e.},
$(\text{clip}_q, \text{win}_q) \rightarrow (\text{clip}_n, \text{win}_n)$, together with the similarity score.
This bookkeeping preserves the exact temporal correspondence that triggered the match.

Although retrieval is performed on windows, these recorded window matches naturally induce clip-level source--target candidates for annotation (Fig.~\ref{fig:data_pipeline}-(6)):
we collect clip pairs $(\text{clip}_q, \text{clip}_n)$ that are supported by high-similarity window matches, and keep the associated matched window indices as evidence for annotators.
In practice, such candidates often share a similar motion backbone for at least one participant but differ in interaction semantics, providing a natural setting for instruction-driven editing: \emph{edit the change, preserve the rest}.

\paragraph{Edit-instruction writing.}
Given a source--target candidate pair, annotators write a free-form edit instruction describing how to transform the source interaction into the target interaction (Fig.~\ref{fig:data_pipeline}-(7)).
Instructions are required to be minimal and actionable: they should describe only the necessary modifications, avoid copying a full target description, and explicitly preserve non-edited content consistent with the source.

\paragraph{Quality control and filtering.}
We apply quality control to remove low-quality or ambiguous candidates (Fig.~\ref{fig:data_pipeline}-(8)).
Typical rejection cases include: (i) the difference between source and target is too subtle to be reliably described, (ii) the instruction is unclear or underspecified, or (iii) the retrieved pair is mismatched (\textit{e.g.}, the matched windows do not reflect the claimed semantic change).
After filtering, the remaining triplets form the final dataset consisting of $(\text{source motion}, \text{target motion}, \text{editing instruction})$ examples (Fig.~\ref{fig:data_pipeline}-(9)).

\paragraph{Train--test similarity analysis.}
We further analyze train--test similarity to examine whether the benchmark contains near-duplicate target motions or edit transformations across splits. Specifically, we compute cosine similarity for target motion embeddings and edit-delta embeddings between training and test pairs. Only $0.0250\%$ of target pairs and $0.00075\%$ of edit-delta pairs exceed a cosine similarity of $0.99$. This indicates low train--test similarity, especially at the edit-transformation level, and suggests that the evaluation is not dominated by near-duplicate samples from the training set.

Overall, Fig.~\ref{fig:data_pipeline} illustrates how retrieval-based candidate mining and human instruction writing jointly enable scalable construction of a dual-person motion editing dataset.
Window-level retrieval provides semantically related candidates with localized temporal correspondences, while human annotation supplies explicit supervision for instruction-following editing models.

\section{Additional Loss Definitions}
\label{sec:appendix_losses}

This appendix provides the mathematical definitions of the geometric and interaction losses used in our objective (Sec.~4.6).
We follow the standard practice in interactive motion diffusion models (\textit{e.g.}, InterGen-style objectives) and report the loss forms here for completeness.

\paragraph{Notation.}
We denote the ground-truth clean two-person motion as $\mathbf{x}_0=(\mathbf{x}_0^{A},\mathbf{x}_0^{B})$ and the denoiser prediction (START\_X) as $\hat{\mathbf{x}}_0=(\hat{\mathbf{x}}_0^{A},\hat{\mathbf{x}}_0^{B})$.
Here, $A$ and $B$ denote the two interacting persons ($p\in\{A,B\}$).

\subsection{Diffusion Reconstruction Loss}
\label{sec:appendix_simple_loss}

With START\_X parameterization, we minimize the reconstruction error on $\mathbf{x}_0$:
\begin{equation}
\mathcal{L}_{\text{diff}}
=
\left\|
\mathbf{x}_0 - \hat{\mathbf{x}}_0
\right\|_2^2
=
\sum_{p\in\{A,B\}}
\left\|
\mathbf{x}_0^{p} - \hat{\mathbf{x}}_0^{p}
\right\|_2^2 .
\label{eq:loss_simple}
\end{equation}

\subsection{Geometric Losses (Per Person)}
\label{sec:appendix_geo_losses}

We employ geometric losses commonly used in human motion generation, including the foot contact loss $\mathcal{L}_{\text{foot}}$ and joint velocity loss $\mathcal{L}_{\text{vel}}$, to regularize the generative model and promote physical plausibility and temporal coherence for each individual motion. For details of these standard losses, we refer the reader to MDM~\cite{tevet2023mdm}.
For non-canonical motion representations, we further introduce a bone length loss $\mathcal{L}_{\text{BL}}$ to enforce skeletal consistency in the global joint positions of each person, thereby implicitly preserving the kinematic structure of the human body.

\paragraph{Bone length loss.}
Let $B(\mathbf{x}^{p})$ represent the bone lengths in a pre-defined human body kinematic tree, derived from the global joint positions:
\begin{equation}
\mathcal{L}_{\text{BL}}
=
\sum_{p\in\{A,B\}}
\left\|
B(\mathbf{x}_0^{p}) - B(\hat{\mathbf{x}}_0^{p})
\right\|_2^2 .
\label{eq:loss_bl}
\end{equation}

\subsection{Interaction Losses (Between Persons)}
\label{sec:appendix_inter_losses}
\paragraph{Masked distance-map loss.}
We define the joint distance map of the two persons as $M(\mathbf{x}^A,\mathbf{x}^B)\in\mathbb{R}^{N_j\times N_j}$, and denote by $M_{xz}(\mathbf{x}^A,\mathbf{x}^B)$ its projection onto the XZ-plane. To focus the supervision on relevant close-range interactions, we apply a binary mask based on the projected distance map. The masked distance-map loss is
\begin{equation}
\mathcal{L}_{\text{DM}}
=
\left\|
\big(
M(\hat{\mathbf{x}}_0^{A},\hat{\mathbf{x}}_0^{B})
-
M(\mathbf{x}_0^{A},\mathbf{x}_0^{B})
\big)
\odot
\mathbbm{1}\!\left(
M_{xz}(\mathbf{x}_0^{A},\mathbf{x}_0^{B}) < \tilde{M}
\right)
\right\|_F^2 ,
\label{eq:loss_dm}
\end{equation}
where $\tilde{M}$ is a distance threshold, $\mathbbm{1}(\cdot)$ is an indicator function, $\odot$ denotes the Hadamard product, and $\|\cdot\|_F$ is the Frobenius norm.
\paragraph{Relative orientation loss.}
We penalize mismatched relative orientation between the two people. Let $IK(\cdot)$ denote the inverse kinematics process that outputs joint rotations, and let $O(\cdot,\cdot)$ denote the 2D relative orientation between the two people around the Y-axis. The loss is defined as
\begin{equation}
\mathcal{L}_{\text{RO}}
=
\left\|
O\!\big(IK(\hat{\mathbf{x}}_0^{A}),IK(\hat{\mathbf{x}}_0^{B})\big)
-
O\!\big(IK(\mathbf{x}_0^{A}),IK(\mathbf{x}_0^{B})\big)
\right\|_2^2 .
\label{eq:loss_ro}
\end{equation}

\subsection{Full Objective}
\label{sec:appendix_full_obj}

Then the motion loss can be summarized as:
\begin{equation}
\mathcal{L}_{\text{motion}}
=
\mathcal{L}_{\text{diff}}
+
\lambda_{\text{vel}}\mathcal{L}_{\text{vel}}
+\lambda_{\text{foot}}\mathcal{L}_{\text{foot}}
+\lambda_{\text{BL}}\mathcal{L}_{\text{BL}}
+\lambda_{\text{DM}}\mathcal{L}_{\text{DM}}
+\lambda_{\text{RO}}\mathcal{L}_{\text{RO}}
\label{eq:loss_motion_appendix}
\end{equation}
To balance the contribution of each term, we set
$\lambda_{\text{vel}}=30$,
$\lambda_{\text{foot}}=30$,
$\lambda_{\text{BL}}=10$,
$\lambda_{\text{DM}}=3$,
$\lambda_{\text{RO}}=0.01$ in all experiments.
Finally, the overall training objective (Sec.~4.6) adds our two auxiliary alignment terms:
\begin{equation}
\mathcal{L}_{\text{total}}
=
\mathcal{L}_{\text{motion}}
+
\lambda_{p}\mathcal{L}_{\text{plan}}
+
\lambda_{f}\mathcal{L}_{\text{freq}} .
\label{eq:loss_total_appendix}
\end{equation}
``Plan'' aligns learnable plan tokens to the teacher embedding of the ground-truth target, and ``Freq'' aligns learnable frequency tokens to the DCT band-energy descriptors derived from interaction signals.

\section{Failure Case Analysis}
\begin{figure*}[t]
  \centering
  \setlength{\tabcolsep}{0pt} 
  \renewcommand{\arraystretch}{1.0}

\begin{tabular}{@{}p{0.495\textwidth}@{\hspace{2pt}}p{0.495\textwidth}@{}}
\begin{minipage}[t]{\linewidth}
  \centering
  \begin{overpic}[width=\linewidth]{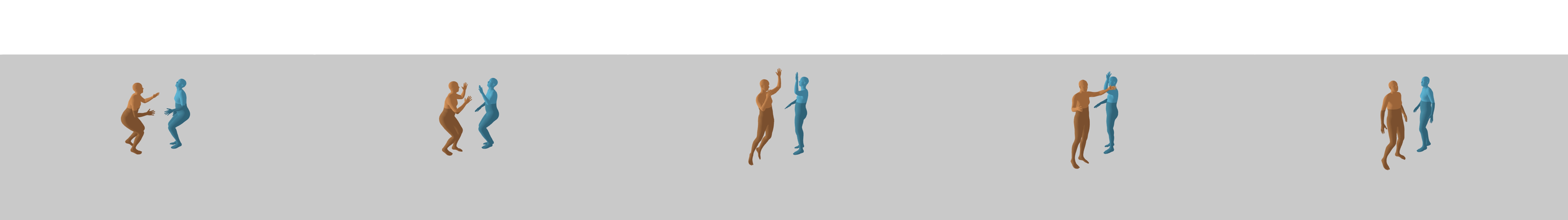}
    \put(1,11){{\scriptsize\bfseries Source}}
  \end{overpic}\\[-1mm]
  \begin{overpic}[width=\linewidth]{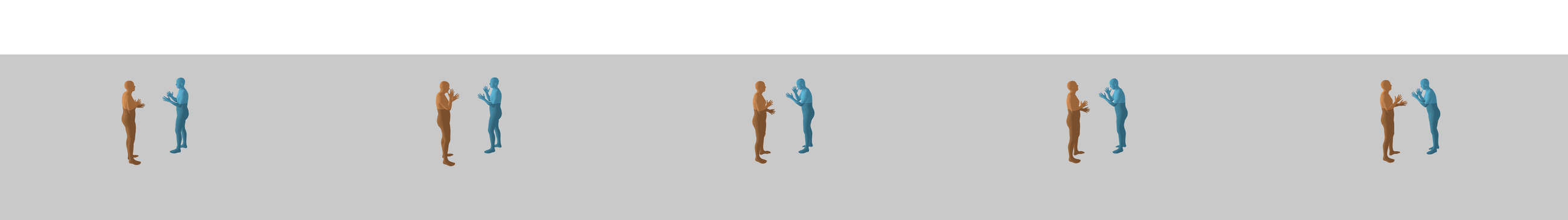}
    \put(1,11){{\scriptsize Ours}}
  \end{overpic}\\[-1mm]
  \begin{overpic}[width=\linewidth]{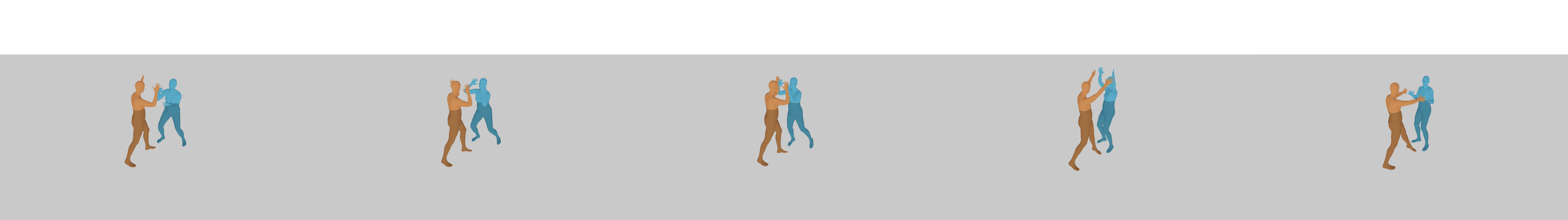}
    \put(1,11){{\scriptsize\bfseries GT Target}}
  \end{overpic}\\[-1mm]
  \medskip
  \parbox{0.98\linewidth}{\centering\small
    \textit{\textcolor{red!70!black}{Clap} with both hands instead of one hand.}
  }
\end{minipage}
&
\begin{minipage}[t]{\linewidth}
  \centering
  \begin{overpic}[width=\linewidth]{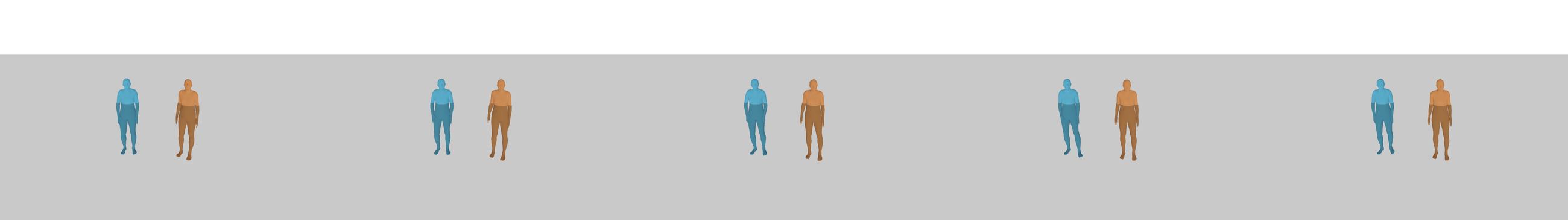}
    \put(1,11){{\scriptsize\bfseries Source}}
  \end{overpic}\\[-1mm]
  \begin{overpic}[width=\linewidth]{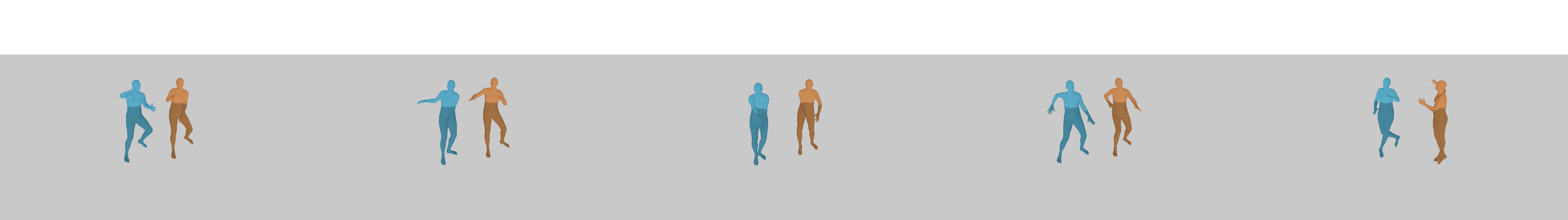}
    \put(1,11){{\scriptsize Ours}}
  \end{overpic}\\[-1mm]
  \begin{overpic}[width=\linewidth]{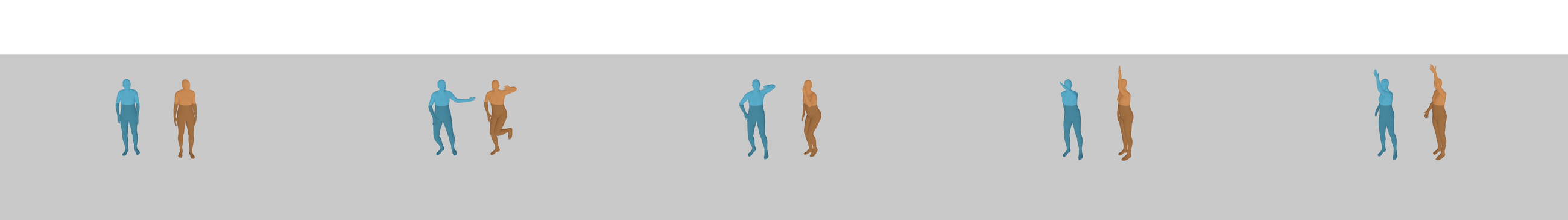}
    \put(1,11){{\scriptsize\bfseries GT Target}}
  \end{overpic}\\[-1mm]
  \medskip
  \parbox{0.98\linewidth}{\centering\small
    \textit{Both individuals dance \textcolor{red!70!black}{shoulder to shoulder} instead of standing still.}
  }
\end{minipage}
\\

  \end{tabular}
  \caption{Failure cases of the proposed method.}
  \label{fig:failure_cases}
\end{figure*}

Fig.~\ref{fig:failure_cases} shows failure cases:
(i) \textit{Ambiguity in gestures:} The model confuses clapping both hands with the other person with clapping with oneself, suggesting challenges in distinguishing subtle gesture variations.
(ii) \textit{Long-horizon relational consistency:} The model struggles to maintain spatial relations in long sequences, as seen in the dance example, where one person eventually drifts. This shows the difficulty in preserving strict inter-person spatial relations during complex, high-motion dynamics over time.

\section{Human Evaluation on Qualitative Comparisons}
\label{sec:human_eval}
Since learned retrieval and FID metrics are proxy measures of editing quality, we conduct a human preference study against the TIMotion baseline on $20$ diverse text prompts, further assessing instruction adherence, source preservation, and interaction realism.
For each prompt, we render the edited two-person motion from both methods with identical visualization settings (camera, duration, frame rate).
To ensure a fair comparison, we anonymize the methods as \textit{A} and \textit{B} and randomize their left/right ordering per sample.

\paragraph{Protocol.}
Participants are shown the source motion (for reference) together with two edited results (A/B).
For each sample, they answer four questions with three options (\textit{A}, \textit{B}, \textit{Tie}):
(i) overall preference,
(ii) instruction adherence,
(iii) source preservation, and
(iv) interaction realism (spatio-temporal coupling).
We collect responses from $10$ participants, and each prompt receives $10$ independent judgments.

\paragraph{Metrics.}
For each criterion, we report the win/tie/lose rates of InterEdit against TIMotion over all judgments.

\paragraph{Evaluation Questions.}

\begin{itemize}
    \item \textbf{Text instruction:} \textit{``<PROMPT>''}
    \item \textbf{Motions:} \textbf{Source}, \textbf{Result A}, \textbf{Result B}
\end{itemize}

\noindent\textbf{Questions.} For each question, participants select one of \{\textit{A}, \textit{B}, \textit{Tie}\}.
\begin{enumerate}
    \item \textbf{Overall preference:} Overall, which result is better?
    \item \textbf{Instruction adherence:} Which result better follows the text instruction?
    \item \textbf{Source preservation:} Which result better preserves the source motion, except for the required edits?
    \item \textbf{Interaction realism:} Which result shows more realistic two-person interaction (relative timing, synchronization, and spatial configuration)?
\end{enumerate}

\paragraph{Results.}
As shown in Table~\ref{tab:human_eval}, InterEdit is preferred over TIMotion across all criteria, with particularly large gains in instruction adherence and interaction realism. This suggests that semantic-aware plan token alignment improves high-level intent following, while interaction-aware frequency token alignment enhances spatio-temporal coupling.

\begin{table}[t]
\centering
\begin{tabular}{lccc}
\toprule
Criterion & Win & Tie & Lose \\
\midrule
Overall preference & 75.5 & 18.0 & 6.5 \\
Instruction adherence & 78.5 & 15.5 & 6.0 \\
Source preservation & 71.0 & 21.0 & 8.0 \\
Interaction realism & 81.0 & 10.5 & 8.5 \\
\bottomrule
\end{tabular}
\caption{Human evaluation on 20 prompts. Win/Tie/Lose rates (\%) of InterEdit vs.\ TIMotion.}
\label{tab:human_eval}
\end{table}

\end{document}